# Align Generative Artificial Intelligence with Human Preferences: A Novel Large Language Model Fine-Tuning Method for Online Review Management

Yanan Wang, Department of Information Systems and Operations Management, The University of Texas at Arlington, yanan.wang@uta.edu, 701 S. Nedderman Drive, Arlington, TX 76019

Yong Ge, Department of Management Information Systems, University of Arizona, yongge@arizona.edu, McClelland Hall, 1130 E Helen St, Tucson, AZ 85721

## Abstract

Online reviews have played a pivotal role in consumers' decision-making processes. Existing research has highlighted the significant impact of managerial review responses on customer relationship management and firm performance. However, a large portion of online reviews remains unaddressed due to the considerable human labor required to respond to the rapid growth of online reviews. While generative AI (especially Large Language Model (LLM)) has achieved remarkable success in a range of tasks, they (i.e., generative AI such as GPT-4) are general-purpose models and may not align well with domain-specific human preferences. To tailor these general generative AI models to domain-specific applications, finetuning is commonly employed. Nevertheless, several challenges persist in finetuning with domain-specific data, including hallucinations, difficulty in representing domain-specific human preferences, and over conservatism in offline policy optimization. To address these challenges, we propose a novel preference finetuning method to align an LLM with domain-specific human preferences for generating online review responses. Specifically, we first identify the source of hallucination and propose an effective context augmentation approach to mitigate the LLM hallucination. To represent human preferences, we propose a novel theory-driven preference finetuning approach that automatically constructs human preference pairs in the online review domain. Additionally, we propose a curriculum learning approach to further enhance preference finetuning. To overcome the challenge of over conservatism in existing offline preference finetuning method, we propose a novel density estimation-based support constraint method to relax the conservatism, and we mathematically prove its superior theoretical guarantees. Extensive evaluations employing objective evaluation metrics, human assessments and qualitative analyses substantiate the superiority of our proposed preference finetuning method. For practical deployment in real-world systems, we recommend two types of hybrid approaches to synergize human and LLM capabilities, which can significantly reduce human labor and time in responding to online reviews.

## 1. Introduction

Online reviews play a crucial role in the consumer decision-making process by providing information regarding the quality of products or services (Chevalier and Mayzlin 2006). Remarkably, nearly 93% of consumers have made purchasing decisions based on online reviews (Howarth 2023). These reviews can be categorized into positive and negative sentiments. Notably, negative reviews can damage a brand's image and propagate negative word-of-mouth, thereby compelling firm managers to address these reviews promptly to resolve consumer complaints. Meanwhile, responding to positive reviews can be beneficial for further enhancing consumer's trust (Deng and Ravichandran 2023). Given the importance of responding to customer reviews, extant research has investigated the impact of human's response on various outcomes, including customer's satisfaction (Gu and Ye 2014), future review volume and valence (Ravichandran and Deng 2023, Deng and Ravichandran 2023), and firm's financial performance (Xie et al. 2017, Kumar et al. 2018). Despite the well-documented benefits of responding to reviews, the process entails considerable financial and time costs for generating human's responses and a substantial portion of online reviews remains unaddressed (Howarth 2023), largely due to the extensive human resources required to manage the rapidly increasing volume of online reviews (Ravichandran and Deng 2023). Therefore, there is an urgent need to automate the review response process to improve efficiency while maintaining effectiveness.

Recently, the development of generative AI, such as large language models (LLM) and large vision models, has marked a revolutionary advancement in the AI field. These generative AI models are trained using massive amount of data and contain an extensive number of parameters, reaching into the billions or trillions. They have demonstrated impressive performance across a wide array of tasks, including chatbot (OpenAI et al. 2024), image generation (OpenAI 2023), and video generation (OpenAI 2024). However, these general-purpose models often lack domain-specific knowledge, which may not align well with domain-specific human preferences in real-world applications. Consequently, fine-tuning is commonly employed to adapt these generative AI models, especially LLMs, to specific domains. Examples include content marketing to create content for website landing pages in search engine optimization (Reisenbichler et al. 2022), medical consultation to provide responses to patient inquires by incorporating medical domain knowledge (W. Wang et al. 2024), customer services to address consumer queries (Gourav 2024), and financial applications to generate financial advices (X. Y. Liu et al. 2023). Particularly, LLM finetuning for domain-specific applications has shown several advantages. (1) Finetuning a pretrained LLM using domain specific data can tailor it to the unique nuances in a business environment (AdVon Commerce 2024). Comparing with prompt engineering methods, LLM finetuning is more effective as it allows deeper integration of domain-specific knowledge and continuous learning from new data (MyScale 2024). (2) Finetuning an open-source LLM (e.g., Llama2) allows the firms to self-host the LLM, thereby saving cost from using closed-source LLM's APIs (e.g., GPT-4) and reducing potential data privacy and security risks. (3) Finetuning an open-source LLM allows us to better customize it for domain-specific applications and tackle some fundamental technical challenges in the LLM literature (e.g., theory-driven preference finetuning and over conservatism in existing offline preference finetuning methods, unfolded in Section 3), which cannot be done with the highly restricted closed-source LLMs (e.g., OpenAI's LLMs). Therefore, in this study we focus on finetuning an open-source LLM to align it with domain-specific human preferences in the review response generation task (i.e., the customer's preferences about which kind of response is better to address the customer review), which is underexplored in current online review management literature.

However, there are several technical challenges in designing a LLM preference finetuning method for review response generation. *First*, when using the collected historical customer reviews and human managerial responses for LLM finetuning, it is observed that LLM-generated responses often contain numerous fabricated facts about customers, firms, and their interactions. These unfaithful facts, termed as "hallucinations" (Ji et al. 2023), can significantly harm customer's satisfactions and damage firm's reputation (Colburn 2024). *Second*, existing preference fine-tuning methods such as Reinforcement Learning from

Human Feedback (RLHF) (Ouyang et al. 2022) and Direct Preference Optimization (DPO) (Rafailov et al. 2023) typically rely on preference pair data, where preferred responses contrast with less preferred ones. However, in the collected review responses data, there is only one human response available for each customer review, making it challenging to represent contrastive preferences for effective LLM preference finetuning. *Third*, preference fine-tuning heavily depends on pre-collected offline data, as online preference optimization with direct human interactions incurs prohibitively high costs (Rafailov et al. 2023). In offline preference optimization, existing methods (e.g., DPO) usually employ a distributional constraint, specifically the Kullback-Leibler divergence constraint, between the learned policy and the offline data distribution. However, the distributional constraint is often overly conservative and unfortunately restricts the learned preference policy to suboptimal offline data, thus degrading the quality of LLM-generated responses.

To address the above challenges, we propose a novel context-augmented theory-driven preference finetuning solution with relaxed conservatism for review response generation. Our solution consists of four technical modules (illustrated in Figure 1) that we design based on data-driven reasoning, relevant IS theories, and machine learning theoretical analyses. In addition to a mathematical proof that validates the superior theoretical guarantees of our technical designs, we evaluate our proposed solution on a hotel review response task against several state-of-the-art baseline methods using both objective evaluation metrics and human assessments, validate and interpret the effectiveness of each specific design through ablation studies and qualitative analyses. The evaluation results demonstrate the superiority of our solution (e.g., 17.0% improvement over the best baseline on an objective evaluation metric and a win rate of 71.8% on human's overall assessment) for review response generation.

## 2. LITERATURE REVIEW

### 2.1 Online Review Response Generation

Given the substantial volume of customer reviews and the critical importance of responding to these reviews (Ravichandran and Deng 2023, Deng and Ravichandran 2023), there has been a growing interest in automating review response generation. Previous studies have primarily concentrated on deep learning methodologies based on the sequence-to-sequence (seq2seq) framework (Sutskever et al. 2014) without incorporating Large Language Models (LLMs) for generating responses (Kew et al. 2020, Farooq et al. 2020). The typical seq2seq model utilizes an encoder to process the review text, followed by a decoder to generate the response. Zhang et al. (2023) introduced a transformer-based seq2seq model that incorporates app category and user rating features to better understand review content and user sentiment. Pretrained language models like BERT and RoBERTa have also been applied to app review response generation, demonstrating robust performance relative to the training data volume. However, these approaches typically rely on seq2seq models trained from scratch or utilize smaller pretrained language models, which have limited capacity compared to more advanced large language models like GPT-4 (OpenAI et al. 2024) or LLaMA2 (Touvron et al. 2023) that are pretrained on extensive datasets. Due to the indisputable inferiority, we exclude these methods in our selected baseline methods in the evaluations. Concurrent with our work, Azov et al. (2024) employed a self-refine approach (Madaan et al. 2023) for iterative prompt engineering to optimize the prompt for generating responses to app reviews. Our proposed method is fundamentally different from theirs because we focus on fine-tuning LLM rather than prompting engineering.

## 2.2 LLM Finetuning

There are three representative types of LLM finetuning techniques: supervised finetuning (SFT), reinforcement learning from human feedback (RLHF), and direct preference optimization (DPO). The first type, supervised fine-tuning, is primarily focused on instruction tuning. The latter two methods fall under the category of preference fine-tuning, which aims to align LLMs with observed human preferences.

### 2.2.1 Supervised Finetuning (SFT)

SFT of LLM involves training models using labeled input-output pairs for end-to-end instruction tuning, where the input typically consists of the instruction description and related task information, and the output is the designated response. As one representative method, LIMA (Zhou et al. 2023) is fine-tuned on a LLaMA-65B model with the standard supervised loss with only 1,000 meticulously selected prompts and responses. To mitigate the costs associated with collecting human-written instruction data. Wang et al. (2024) applied SFT to develop a medical consultation chatbot, utilizing curated instruction data informed by medical domain knowledge, including soft skills and fundamental principles practiced by human doctors. Although

SFT is less computationally demanding, it has been observed that the performance of supervised fine-tuned LLMs consistently falls short of those enhanced via preference finetuning (Ouyang et al. 2022, Touvron et al. 2023, Rafailov et al. 2023, OpenAI et al. 2024).

### 2.2.2 Reinforcement Learning from Human Feedback (RLHF)

To address the limitations of SFT methods, RLHF has been proposed and implemented in many powerful LLMs such as GPT-4 (OpenAI et al. 2024), Llama2 (Touvron et al. 2023). One of the most representative RLHF methods is instructGPT (Ouyang et al. 2022), which consists of three training steps. The first step is to collect human demonstration or instruction data and fine-tune an LLM using supervised learning (SFT stage). The second step involves gathering a dataset of output comparisons, wherein labelers identify their preferred outputs for specific inputs. This facilitates the training of a reward model designed to predict the human-preferred output, thus modeling human preferences (reward modeling stage). The third step is to fine-tune the LLM obtained in step 1 to optimize the reward using reinforcement learning (RL) algorithms such as Proximal Policy Optimization (PPO) (Schulman et al. 2017) (RL training stage). Furthermore, the second and third steps can be iteratively executed by collecting fresh human feedback online using the RL fine-tuned LLM post-third step. However, RLHF training involves multiple LLMs, such as an LLM for reward modeling and another for value modeling during the RL training phase, which imposes a significant computational burden and demands extensive computing resources (Rafailov et al. 2023). Moreover, the PPO optimization used in the RL training phase is notably sensitive to hyperparameter settings, increasing training instability (Rafailov et al. 2023). Additionally, as RLHF explicitly constructs a reward model, imperfect reward modeling can lead to misalignment of the fine-tuned LLM with human preference (Chaudhari et al. 2024).

### 2.2.3 Direct Preference Optimization (DPO)

Given the complexity and instability associated with the RLHF pipeline, recent methodologies have aimed to bypass the reward modeling step and directly optimize human preferences. DPO (Rafailov et al. 2023) is a prominent example of such methods and is mathematically demonstrated to be equivalent to the traditional RLHF's learning objective (Rafailov et al. 2023). Unlike RLHF, DPO merges reward modeling and RL upolicy optimization into a singular process in an offline setting by utilizing an analytical mapping between the RL fine-tuned policy and the reward function, thereby eliminating the need for explicit reward modeling. Following DPO, various alternative preference optimization methods have recently been proposed. For instance, Preference Optimization with Identity Mapping (IPO) modifies the logistic regression loss used in DPO with a squared regression loss, aiming to mitigate potential overfitting to deterministic preference samples in the dataset (Azar et al. 2024). SLiC-HF employs hinge loss to maximize the margin between the likelihoods of the preferred and less preferred samples (Zhao et al. 2023). Ethayarajh et al. (2024) proposed to maximize the utility of model outputs, diverging from the traditional focus on maximizing log-likelihood of preferences, and instead drawing on the human utility function described in Kahneman and Tversky's prospect theory (Tversky and Kahneman 1992).

Similar to RLHF and DPO methods, our solution initiates with supervised finetuning (SFT) of the LLM and subsequently employs preference fine-tuning to further align the initialized LLM with human preferences. Given the computational efficiency, stable performance, and superior outcomes exhibited by recently developed LLMs (Ivison et al. 2023, Abdin et al. 2024), we develop a novel solution for automated review response based on the DPO technique (Rafailov et al. 2023).

## 2.3 LLM Hallucination

Hallucination has been identified as a significant challenge in developing reliable LLMs for practical applications (Huang et al. 2023, Maleki et al. 2024, Venkit et al. 2024). Huang et al. (2023) delineated three primary sources of hallucination throughout the LLM capability acquisition process: the data used for training LLM (Singhal et al. 2023), the training process itself (Schulman 2023), and the inference or generation process (Y. Chen et al. 2023). For instance, data-induced hallucination may arise from the absence of relevant knowledge specific to a domain application (Y. Zhang et al. 2023). In response to these causes, three types of hallucination mitigation methods have been developed (Huang et al. 2023). As one representative method, Retrieval Augmented Generation (RAG) is widely employed to mitigate hallucination by bridging knowledge gaps with external knowledge, particularly when domain-specific information is absent (Gao et al. 2024). However, RAG-based methods are not applicable to our setting due to the unavailability of specific information related to customers or hotels. Within the setting of this study, we identify the source of hallucination is the context information gap between customer reviews and human responses in the training data used for finetuning a LLM. We then propose an effective method to extract objective domain-specific facts from human responses in the training data and use them to augment LLM's inputs. Such data augmentation bridges the context information gap, thus mitigating the hallucinations in LLM-generated responses.

## 3. CONTEXT-AUGMENTED THEORY-DRIVEN PREFERENCE FINETUNIING WITH RELAXED CONSERVATISM

### 3.1 Overview

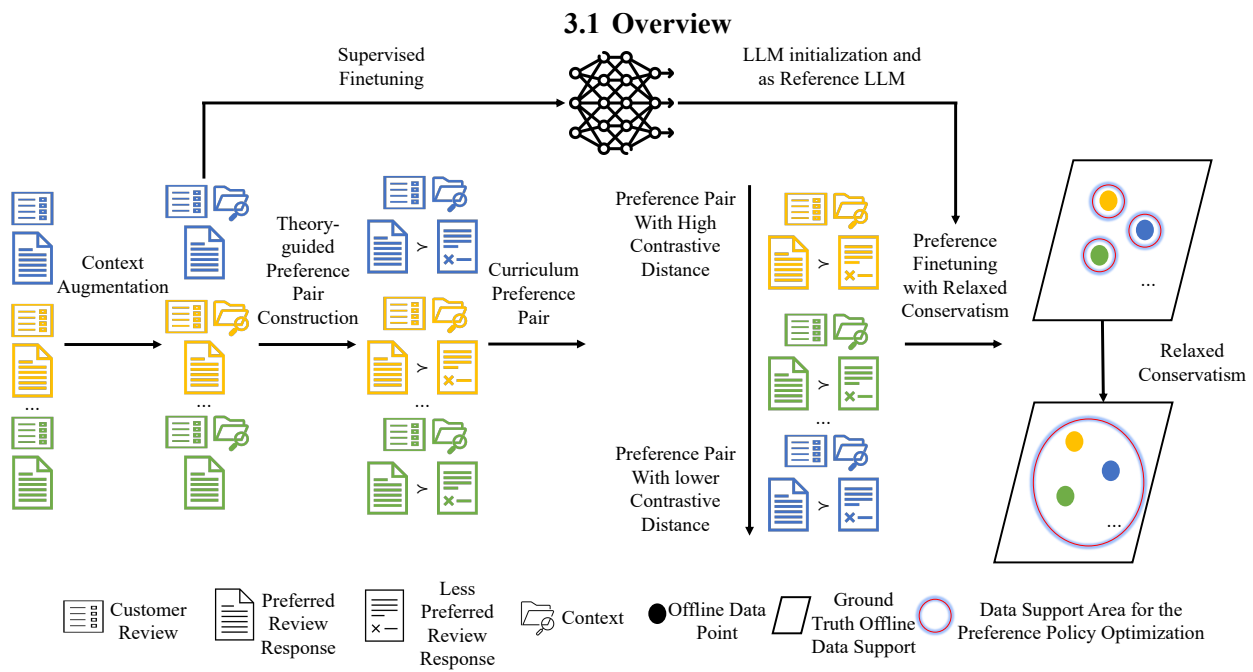

Figure 1. Overview of Our Proposed Context-augmented Theory-driven Preference Finetuning with Relaxed Conservatism Method. Human responses are considered as preferred review responses.

Given the training data $\{(x, y)\}$, where $x$ represents the customer review, $y$ is the corresponding human-written response, the objective of our study is to develop a fine-tuning method for large language models (LLMs) that enables them to automatically generate responses. The goal is for the generated responses to align with customer preferences or maximize customer satisfaction in addressing the review. Figure 1 provides an overview of the proposed LLM preference learning method for online review management, which consists of four design modules. Utilizing the historical customer reviews and human managerial responses to finetune large language models (LLM) may yield unfaithful facts, termed as "hallucinations," in

LLM-generated responses. One significant contributor to such hallucinations is the source-reference divergence from data (Ji et al. 2023). Therefore, as the *first* design module, our method augments the historical review-response data with additional context facts to bridge the knowledge gap between reviews and responses, thereby alleviating the hallucinations issue. In line with common practices in LLM fine-tuning, we initially train a LLM using the SFT technique and the context-augmented data. The initialized LLM through the SFT process will serve as the initialization and the reference model (detailed in Section 3.3.3) during the subsequent preference fine-tuning phase.

LLM fine-tuning methods for preference alignment such as RLHF (Ouyang et al. 2022) or DPO (Rafailov et al. 2023) typically require training datasets composed of pairwise contrastive preferences (i.e., a preferred response versus a less preferred response for one given prompt). However, in the context of online review management (i.e., the focus of our study), the dataset predominantly contains only the preferred human responses, lacking explicitly contrastive preferences. To address this challenge, the *second* module of our method constructs contrastive preference pairs, guided by justice theory, principles of service recovery, type of review, and review sidedness, as delineated in the information systems (IS) literature (Ravichandran and Deng 2023, Deng and Ravichandran 2023). The construction is performed in an automatic way through our designed prompt engineering methods with GPT-4, where we design a novel constrained prompt engineering method for generating less preferred responses to negative reviews. The constructed contrastive pairs of responses are used for fine-tuning the initialized LLM, facilitating the alignment of human preferences in LLM-generated responses.

The degree of contrast between the preferred and less preferred responses varies across different pairs constructed in the second module. This variance poses a challenge for preference learning during LLM finetuning (Amini et al. 2024), particularly when the contrastive difference within a preference pair is minimal, making it difficult to discriminate between the preferred and less preferred responses. To cope this challenge, as the *third* module of our method, we propose a curriculum learning (Bengio et al. 2009) approach to achieve a more effective preference fine-tuning process, where we order the preference pairs by their contrastive differences, ranging from easy (large contrastive distance) to difficult (small contrastive distance) pairs, and use them sequentially to fine-tune the initialized LLM through SFT.

Finally, preference learning very much relies on pre-collected offline data because online preference optimization through direct human interaction involves extremely high cost (Rafailov et al. 2023). Existing offline preference optimization methods (e.g., Direct Preference Optimization (DPO)) for LLM fine-tuning incorporate distributional constraints—specifically the Kullback-Leibler divergence constraint—between the learned policy and the distribution of the offline data (modeled by the reference LLM policy learned through the SFT). However, these constraints can be overly conservative and thus degrade the performance of the learned preference policy due to the suboptimality of offline data (Kumar 2019, Levine et al. 2020, Singh et al. 2023). To overcome this fundamental LLM finetuning challenge, in the *fourth* module we propose a novel density estimation-based support constraint approach to relax the conservatism during the preference policy optimization, while still maintaining adherence to the offline data support; and we mathematically prove its superior theoretical guarantee compared to DPO.

We summarize the research challenges, research gaps and our technical contributions in Table 1.

Table 1. Summary of the research challenges, research gaps and our technical contributions.

| Research Challenges | Research Gaps | Technical Contributions |
|---|---|---|
| Hallucinations in LLM finetuning for review response generation | Existing literature fails to identify the source of hallucination in the review response generation setting and the RAG-based hallucination mitigation methods (Gao et al. 2024) typically rely on external information, which are usually not publicly available. | During LLM finetuning, for each review-response pair in the training data, we extract the faithful information from the response and then augment the review with this extracted faithful information. The proposed hallucination mitigation method can be generalized to other domain-specific LLM finetuning. |
| Represent human preferences for the generated managerial responses for LLM preference finetuning | Existing successful preference finetuning methods (Ouyang et al. 2022, Rafailov et al. 2023) require pairwise preference data. However, in the review response generation setting, it lacks preference pair data and it's unknown how to represent human preferences. | We propose an automatic preference data construction method grounded in theories in the IS literature and use it for LLM preference finetuning. The proposed method enables scalable preference alignment in the online review response generation setting. |
| The degree of contrast between the preferred and less preferred responses varies across different preference pairs and impact the learning difficulty level of each | Existing preference finetuning methods (Ouyang et al. 2022, Rafailov et al. 2023) typically ignore the contrastive difference among preference pairs, where the preference pairs are often randomly sampled from the training dataset. | We propose a curriculum learning approach by ordering the training preference pair samples from the largest to the smallest contrastive distance to improve the effectiveness of preference finetuning. The proposed method can be |

| preference pair sample for LLM preference finetuning | | combined with other general preference finetuning methods. |
|---|---|---|
| Excessive conservatism in offline preference optimization | Existing offline preference finetuning methods (Rafailov et al. 2023) impose overly conservative constraints that confine the learned policy strictly to the potential suboptimal offline training data. | We first provide a mathematical analysis to reason the cause of the excessive conservatism. Then, we propose to relax conservatism with a designed density estimation-based support constraint approach. We also provide theoretical analysis to demonstrate the superiority of this proposed approach. The proposed approach can be integrated with general offline preference finetuning methods. |

### 3.2 Context Augmentation to Mitigate LLM Hallucination

**Customer Review:** Substandard ---SEP--- This motel is in a lovely location with trees and gardens to see and enjoy; it's close to bush-walks and several shops. ⋯ This motel has no reception office nor staff on site!!! On arrival one has to 'phone a nearby shop to arrange being admitted; consequently we had clients knocking on our door and window thinking our ⋯

**Context Information Gap:**
**1**. The motel provides a kettle, microwave, toaster, and fridge as part of its amenities. **2**. The motel is planning to create a BBQ area for guests. **3**. The fridge is placed inside a large under bench cabinet with plenty of airspace to limit motor noise. ⋯ **7**. At the time of the customer's stay, the motel was fully booked, and all cheaper rooms had been taken.

**Human Managerial Response:** Thank you for your review as all feedback is important to us and we will definitely take your comments on board. ⋯ We don't live on site, but we are always accessible by phone, and we are only 5-10 minutes away if needed. At the time you stayed, we were fully booked, and unfortunately, all our cheaper rooms had been taken ⋯

Figure 2. An example of hotel's customer review, human managerial response, and context information gap. The complete text is provided in Online Appendix A-1.

As shown in Section 3.1, our proposed method begins with supervised fine-tuning (SFT) of a LLM using historical review-response data. However, directly utilizing the collected pairs of customer review $x$ and human's response $y$ to fine-tune the LLM will cause it to hallucinate. In the following, we first explain the source of hallucination in our review-response setting and then propose a simple yet effective hallucination mitigation strategy for the SFT of LLM.

Given a customer review $x$, human's managerial response $y$ may include some faithful facts to address the review. These facts play as important context information in the response but are absent in the customer review. In a collected set of reviews and responses (i.e., pairs of $x$ and $y$), there is often such context information gap between each customer review and its response. An illustration example extracted from a real-world dataset is shown in Figure 2. This kind of context information is very specific to a particular customer, a target hotel and the interaction between the customer and the hotel. During the supervised fine-tuning stage, the LLM learns to mimic the human response $y$ given customer review $x$, despite the absence of the context. This supervised fine-tuning process can be viewed as behavior cloning in the reinforcement learning (RL) literature (Pomerleau 1991). Simply cloning human behavior (i.e., human response) without addressing the context information gap between input (review) and output (response) compels the LLM to generate fabricated details about customer, product/service (e.g., hotel), and their interactions, a process known as hallucination (Ji et al. 2023). Similar phenomena of LLMs have been reported in the literature (Schulman, 2023). Furthermore, the hallucinations are likely to be carried over to the LLM during the preference fine-tuning stage.

To mitigate the hallucination induced by the context information gap, we propose to augment the review and response data (i.e., pairs of $x$ and $y$) with additional context information. Ideally, relevant stakeholders may be able to supply context information related to the products or services that are reviewed by customers (e.g., Hilton can provide context information such as management policy for a specific Hilton hotel). However, such context information is often not publicly accessible, and our collected data only contain customer reviews and human responses (details available in Section 4.1). To tackle this challenge, we extract the context information from human responses by utilizing GPT-4. The specific prompt designed for the extraction through GPT-4 is shown in the Online Appendix A-2. An example output using the designed prompt is shown in Online Appendix A-3. For each pair of review and response, $(x, y)$, the extracted context information from the response is denoted as $c$. The collected set of reviews and responses, denoted as $\{(x, y)\}$, is then augmented with the context information, denoted as $\{(x, c, y)\}$. The context-augmented data are then used for the SFT process in our proposed method. The SFT process is essentially to train a LLM to predict the next token in the response (i.e., $y$) based on the given input (i.e., $x\ and\ c$), which follows the standard supervised learning paradigm for next-token prediction (Ouyang et al. 2022). By incorporating the augmented context $c$ into the SFT, we bridge the context information gap and the LLM is trained not to hallucinate unknown facts, thereby substantially reducing the occurrence of hallucinations in LLM-generated content. The specific format of the input, along with an example, is demonstrated in Online Appendix A-4, where it can be seen that we also include a simple prompt instruction as additional input. The same input format of training data will be also used in the preference finetuning stage (details available in Section 3.3.3). We also conduct a human evaluation to assess the quality of extracted context information in Online Appendix A-22 and the evaluation results demonstrate high quality of the extracted context information.

*Remarks:* The context information is only augmented for training data, and thus there is no data leakage issue during the test phase. Our proposed simple yet effective hallucination mitigation method can be easily adapted for other applications. When the firm's internal documents are available, our proposed mitigation method can be seamlessly integrated with a retrieval model that retrieves relevant context information.

### 3.3 Theory-driven Preference Finetuning

Although the SFT process discussed in Section 3.2 initializes a LLM, it is not purposely trained to distinguish preferred responses from less preferred ones. To this end, we develop a preference fine-tuning approach to further train the initialized LLM to generate human-preferred review responses. In the literature, there are two classes of LLM preference finetuning techniques: reinforcement learning from human feedback (RLHF) (Ouyang et al. 2022) and direct preference optimization (DPO) (Rafailov et al. 2023), both of which have shown great success in a wide range of tasks (Ivison et al. 2023, OpenAI et al. 2024). Nevertheless, they generally require pairwise preference data—consisting of both a preferred and a less preferred response to a given prompt—for fine-tuning. In our review-response setting, such preference pairs data are typically unavailable. To address the challenge, first we draw on justice theory, principles of service recovery, type of review, review sidedness theories and develop automated methods to construct preference pairs for both negative and positive customer reviews (see Sections 3.3.1 and 3.3.2). Then, we introduce a preference finetuning approach based on DPO that maximizes the contrastive distance between the preferred and less preferred responses (see Section 3.3.3). To further enhance its performance, we propose a curriculum learning design and integrate it into our preference finetuning approach (see Section 3.3.4). Our approach systematically organizes the learning process, starting with easy distinctions and progressively advancing to difficult differentiations between responses, thereby effectively enhancing its ability to learn nuanced human preferences in review responses.

**3.3.1 Theory-driven Automatic Preference Pair Construction for Negative Customer Review**

Building upon justice theory (Tax et al. 1998) and service recovery literature (Miller et al. 2000), we develop a framework to construct a less preferred response for each negative customer review, and the observed human response is considered as a preferred response. As the construction of less preferred responses is automated, our approach enables scalable alignment of LLM to human preference with the theoretical principles-guided framework.

Justice theory provides a theoretical framework to elucidate customer complaint behaviors through the lens of perceived justice. Specifically, there are three dimensions of justice: distributive justice, procedural justice and interactional justice (Tax et al. 1998). In the context of customer complaints, distributive justice pertains to the perceived fairness of an outcome. Customers may, for instance, express dissatisfaction if they perceive discrimination (equality), believe they have not received what they deserve (equity), or feel their specific needs have not been met (need) (Tax et al. 1998). Procedural justice refers to the perceived fairness of the procedure where the outcome is produced. Complaints in this dimension may arise from perceptions of undue delay (speed), a lack of adaptability of procedures (flexibility), or difficulties in engaging with the process (accessibility) (Tax et al. 1998). Interactional justice focuses on the fairness of interpersonal treatment during the enactment of processes. Customer complaints regarding interactional justice may focus on the perceived lack of courteous behavior (politeness), insufficient effort in resolving issues (effort), and the absence of caring and individual attention (empathy) (Tax et al. 1998). Recently, Ravichandran & Deng (2023) have applied justice theory to classify customer complaints in online reviews, a methodology we also adopt for categorizing types of customer review in our research.

Service recovery refers to the action taken by a service provider to address customer complaints regarding a perceived service failure (Gronroos 1988). Ravichandran & Deng (2023) synthesize existing studies on effective managerial response strategies for traditional offline service recovery (Karatepe 2006) and those for online forums (Strauss and Hill 2001). Their synthesized managerial response strategies are further classified into rational and emotional ones. Rational response strategies aim to appeal to customer's logic by delivering information in a clear and objective manner (Claeys and Cauberghe 2014). Conversely, emotional response strategies engage customer's emotions, utilizing dramatic elements and integrating subjective and evaluative content (Claeys and Cauberghe 2014). Following the typology used by Ravichandran & Deng (2023), rational cues include actions such as explanation, redress, facilitation, and reinforcement. Emotional cues, on the other hand, comprise apology, appreciation, attentiveness and encouragement.

Table 2. A theory-driven preference pair construction framework for negative customer review.

| Negative Review Type | Description of Negative Review Type | Preferred Response | Criteria of Less Preferred Response |
|---|---|---|---|
| Negative Review Type 1 | Contain more complaints about procedural unfairness than interactional unfairness | Human Response | Managerial responses with more emotional cues than rational cues |
| Negative Review Type2 | Contain more complaints about interactional unfairness than procedural unfairness | Human Response | Managerial responses with more rational cues than emotional cues |
| Negative Review Type 3 | Contain equal number of complaints about interactional unfairness and procedural unfairness | Human Response | Managerial responses with only emotional cues or managerial responses with only rational cues |
| Negative Review Type 4 | Contain only complaints about distributive unfairness | Human Response | N/A |

   Given the diversity of customer complaint types, different combinations of rational and emotional cues are deemed appropriate for managerial responses (Ravichandran and Deng 2023). The findings in Ravichandran & Deng (2023) suggest that when a negative customer review contains more complaints about procedural unfairness (injustice), managerial responses with more rational cues are appropriate, and managerial responses with more emotional cues are appropriate when the negative customer review contains more complaints about interactional unfairness (injustice). Building on these insights, we develop a framework for automatically constructing less preferred responses, detailed in Table 2. Specifically, the construction of less preferred responses for negative review Type 1 and 2 is based on the opposite of the findings in Ravichandran & Deng (2023); For negative reviews that contain an equal number of complaints about procedural and interactional unfairness (Type 3), responses that exclusively employ rational or emotional cues are considered less preferred, as they fail to address either aspect of perceived unfairness adequately; For negative reviews that solely address distributive unfairness (Type 4), we refrain from constructing a less preferred response due to inconclusive hypothesis test results on distributive unfairness noted in Ravichandran & Deng (2023). In addition, considering the significant economic benefit of responding to customer reviews (Ravichandran and Deng 2023), especially customer complaints, the firm managers spare no effort to improve review responses' quality. Therefore, we consider the observed human response as the preferred one to a customer review. Note that we still observe some low-quality human responses, and we apply some criterion to select the high-quality ones (details available in Section 4.1).

Guided by the proposed framework shown in Table 2, we realize the automatic generation of preference pairs through the following technical procedures. We first design a prompt engineering approach to automatically classify each customer review into one of the four negative review types, and then propose another prompt engineering approach to automatically generate the less preferred response for each customer review. For a given customer review, we calculate its complaint scores for each of the three types of unfairness: distributive unfairness, procedural unfairness and interactional unfairness, denoted as $(du, pu, iu)$. $du$ is the summation of values of the three attributes for the distributive unfairness: equality, equity and need, where each value of the attribute is either 0 or 1 depending on whether this customer review contains complaints about this attribute. Similarly, $pu$ adds the values of speed, flexibility and accessibility, and $iu$ sums the values of politeness, effort and empathy. The value of $du, pu,$ and $iu$ is in the range of $[0, 3]$. As GPT-4 has demonstrated superior performance in various natural language processing tasks (OpenAI et al. 2024), we prompt GPT-4 using the survey items developed by Ravichandran & Deng (2023) to automatically determine the value for each of the aforementioned nine attributes (e.g., equality and speed). The detailed prompt design is shown in Online Appendix A-5. Finally, the classification of the negative review type is determined according to Table 2 by comparing the values in $(du, pu, iu)$.

With the classified negative review type for each customer review, we generate less preferred responses by instructing GPT-4 to produce responses that adhere to the criteria for less preferred responses as outlined in Table 2. The detailed categorization of four types of rational and emotional cues outlined in Ravichandran & Deng (2023) are shown in Online Appendix A-6. However, we found it challenging to directly prompt GPT-4 to meet the criteria for the less preferred response generation, where the quantity relation between the rational and emotional cues in GPT-4's outputs often violates the given criteria in Table 2. To address this challenge, we design a novel constrained prompt engineering method. For each negative review type, we first establish a valid set $\{(n_r, n_e) | 0 \leq n_r \leq 4, 0 \leq n_e \leq 4\}$ that satisfies the criteria for the less preferred response, where $n_r$ and $n_e$ represents the numbers of rational and emotional cues respectively. With the predefined valid set, we then randomly sample one pair of $(n_r, n_e)$ for negative review type 1 and 2, and two pairs of $(n_r, n_e)$ for negative review type 3. Subsequently, we randomly select $n_r$ rational cues and $n_e$ emotional cues from the rational and emotional cues set listed in Table A-6-1 in Online Appendix A-6. Finally, we use these sampled cues to prompt GPT-4 to generate a less preferred response by employing a designed prompt template outlined in Online Appendix A-7. This constrained prompt engineering approach enables the generated less preferred response to meet the criteria for each negative review type.

### 3.3.2 Theory-driven Automatic Preference Pair Construction for Positive Customer Review

Table 3. A theory-driven preference pair construction framework for positive customer review.

| Positive Review Type | Description of Positive Review Type | Preferred Response | Less Preferred Response |
|---|---|---|---|
| Positive Review Type 1 | One-sided instrumental review | Human Response | Tailored response |
| Positive Review Type 2 | One-sided affective review | Human Response | Template Response |
| Positive Review Type 3 | Two-sided instrumental review | Human Response | Template response |
| Positive Review Type 4 | Two-sided affective review | Human Response | Tailored response |
| Positive Review Type 5 | Mixed review | N/A | N/A |

Guided by a theoretical principle that was developed by Deng & Ravichandran (2023), we introduce another theory-driven framework for automatically constructing less preferred responses for positive reviews. Positive reviews are classified by type (instrumental or affective) and sidedness (one-sided or two-sided) (Deng and Ravichandran 2023). Instrumental positive reviews provide objective descriptions of product attributes, while affective positive reviews reflect consumer's subjective experiences. A one-sided review exclusively highlights positive aspects, whereas a two-sided review encompasses both positive and negative elements. Deng & Ravichandran (2023) theorize the effectiveness of tailoring responses to each review type in influencing future review valence. Based on their insights, we propose a framework for constructing less preferred responses to positive reviews as outlined in Table 3. The rationale behind the constructed less preferred response for each review type is as follows. (1) For the two-sided instrumental review (Type 3), a template response that fails to address the issues may suggest managerial disregard for customer feedback. (2) Tailoring a response to a one-sided instrumental review (Type 1) by emphasizing positive product attributes could be perceived as overly promotional, potentially raising questions about the sincerity of the response. (3) for a two-sided affective review (Type 4), the tailored response may incorrectly emphasize the ambiguous negative aspects of the review. (4) As a one-sided affective review (Type 2) typically presents less ambiguity, a template response could indicate inattentiveness to customer's positive emotions. Given the demonstrated negative impact of responding to mixed reviews (containing both instrumental and affective cues) on future review valence (Deng and Ravichandran 2023), we exclude the response generation for this type of review (Type 5).

Following the theoretical framework in Table 3, we also develop prompt engineering methods to realize the automated construction of preference pairs for positive reviews. We first classify positive reviews into the five types using GPT-4. The designed prompt for the review classification, which is inspired by the survey items in Deng & Ravichandran (2023), is shown in Online Appendix A-8. Subsequently, for each class of positive reviews, we develop another prompting method to automatically generate less preferred responses by instructing GPT-4 to produce either a template or a tailored response according to Table 3. The designed prompt for the generation is shown in Online Appendix A-9. We also conduct a human evaluation to assess the quality of generated less preferred response for both negative and positive reviews in Online Appendix A-23 and the evaluation results demonstrate high quality of the generated less preferred response.

### 3.3.3 Preference Finetuning

With the constructed contrastive preference pairs for both negative and positive reviews, preference learning is to use them to further fine-tune the SFT-initialized LLM, enabling it to generate review responses that align with human preferences and meet customer satisfaction. As introduced in Section 2.2, RLHF (Ouyang et al. 2022) and DPO (Rafailov et al. 2023) are two representative classes of LLM finetuning methods for preference optimization. Rafailov et al. (2023) has demonstrated that DPO possesses mathematical equivalence to the KL-constrained version of RLHF. Given its simplicity and demonstrated effectiveness, we have selected DPO as the preference learning method in this paper. The objective of our DPO-based preference learning is formalized as:

$$\max_{\pi_\theta} \mathbb{E}_{(x,c,y_w,w_l)\sim \mathcal{D}} \left[ \log \sigma \left( \beta \log \frac{\pi_\theta(y_w|x,c)}{\pi_{\text{ref}}(y_w|x,c)} - \beta \log \frac{\pi_\theta(y_l|x,c)}{\pi_{\text{ref}}(y_l|x,c)} \right) \right] \stackrel{\text{def}}{=} \mathcal{J}_{pl}, \quad (1)$$

where $x$ is the customer review; $c$ is the augmented context as discussed in Section 3.2; $y_w$ is the preferred response (i.e., human response); $y_l$ is the less preferred response; $\sigma$ is the logistic function. $\pi_{\text{ref}}$ denotes the reference model/policy that is obtained from the SFT stage, and it is fixed during the preference finetuning process. $\pi_\theta$ denotes the learned LLM preference policy, and it is initialized from the SFT-trained LLM with data $\{(x, c, y_w)\}$. $\beta$ is the hyper-parameter to control the deviation of $\pi_\theta$ from $\pi_{\text{ref}}$. The context-augmented data (i.e., $\{(x, c, y)\}$ constructed in Section 3.2) is also used here for the preference finetuning process.

### 3.3.4 Curriculum Learning For Enhancing Preference Finetuning

During the preference finetuning process, the LLM is trained to increase the likelihood of generating preferred responses while simultaneously decreasing the likelihood for less preferred responses. In the conventional preference fine-tuning process, preference pairs are often randomly sampled from the training dataset to train the LLM, which may not yield the most effective training outcome. Intuitively, when the contrastive distance between the preferred response and the less preferred one in a preference pair is large, it is easy for the LLM to learn to distinguish both. Conversely, when the contrastive distance is small, it becomes challenging for the LLM to learn to differentiate between the preferred and less preferred responses. Therefore, ordering the preference pairs from least to most challenging and using them sequentially to train the LLM may render the fine-tuning process more effective. This intuition aligns with the principle of curriculum learning (Bengio et al. 2009), which involves organizing training samples from easy to hard to improve the training efficacy of machine learning (a detailed literature review on curriculum learning in Online Appendix A-10). However, current LLM finetuning methods largely ignore the impact of training sample order and it is still underexplored how to measure the easy/hard samples for preference finetuning.

To fill this research gap, we propose a curriculum learning approach for enhancing LLM fine-tuning. We measure the data sample difficulty level based on the contrastive distance of a constructed preference pair, and a larger contrastive distance suggests a lower difficulty level for one data sample. We then arrange the constructed preference pairs from the largest to the smallest contrastive distance and use them sequentially to fine-tune the LLM. Specifically, the contrastive distance for the $i$-th preference pair $(x_i, c_i, y_{w,i}, w_{l,i})$ is defined as the likelihood difference between the preferred and less preferred responses, denoted as $prefDist_i = \pi_{ref}(y_w|x,c) - \pi_{ref}(y_l|x,c)$. The training data samples (i.e., preference pairs) are ordered and used in a descending order of $prefDist_i$, thereby aligning the training process with a gradually increasing level of difficulty and ensuring a more structured and effective training experience for the LLM.

### 3.4 Relaxed Conservatism in Offline Preference Optimization

Due to the significant expenses associated with online preference optimization that involves collecting human feedback through direct interactions with users (Rafailov et al. 2023), our preference learning is optimized using the pre-collected offline data (i.e., the context-augmented historical reviews/responses and the constructed preference pairs based on them). However, conventional methods for offline preference optimization tend to impose overly conservative constraints that confine the learned policy strictly to offline training data (Singh et al. 2023). This overly restrictive constraint can potentially undermine the performance of the learned preference policy due to the suboptimality of offline training data, thus degrading the quality of the generated responses by the finetuned LLM. To address this methodological challenge, we develop a more flexible approach by moderating the level of conservatism during the preference policy optimization process, while still ensuring adherence to the support of the offline data.

#### 3.4.1 Theoretical Analysis of Excessive Conservatism in Offline Preference Optimization

Here we provide a theoretical analysis on what causes the excessive conservatism in the DPO-based offline preference learning (Rafailov et al. 2023), formalized in Equation 1, based on which we will introduce our novel density estimation-based support constraint approach to relax the conservatism. In the following analysis, we omit the augmented context $c$ in Equation 1 for simplicity.

Equation 1 can be derived from the same reward optimization objective as RLHF (Ouyang et al. 2022):

$$\max_{\pi} \mathbb{E}_{x \sim \mathcal{D}, y \sim \pi}[r(x,y)] - \beta \mathbb{D}_{\mathrm{KL}}[\pi(y|x) \parallel \pi_{\mathrm{ref}}(y|x)], \qquad (2)$$

where $r$ is the reward model in the second stage of RLHF trained using pairwise comparison data; $\mathbb{D}_{\mathrm{KL}}$ represents Kullback–Leibler (KL) divergence; both $\pi_\theta$ and $\pi_{\mathrm{ref}}$ are initialized by the SFT in the first stage of RLHF, with $\pi_{\mathrm{ref}}$ being fixed during the third stage of training. The first term in Equation 2 is the conventional reward maximization objective, while the second term aims to prevent the learned policy from significantly deviating from the reference policy. In the context of offline policy preference optimization, the objective in Equation 2 is essentially a special case of the objective of the offline reinforcement learning algorithm AWR (Peng et al. 2019) with two modifications: 1) the advantage value in AWR is replaced by the reward value in Equation 2; 2) the behavior policy (the policy that generated the offline data) in AWR is replaced by the reference policy $\pi_{\text{ref}}$. And $\pi_{\text{ref}}$ obtained using the SFT on the offline data can be seen as one way to estimate the behavior policy. A significant challenge in reinforcement learning from offline data is the distribution shift (Levine et al. 2020), which refers to the discrepancy between the distribution of the offline data and the distribution induced by the learned policy. Many offline RL methods implement policy constraints to keep the learned policy close to the behavior policy (Levine et al. 2020). The KL-divergence constraint used in AWR is one type of distribution constraint (Levine et al. 2020). It forces the distribution matching between the learned policy and the behavior policy, which can be overly conservative and lead to the suboptimal learned policy when the offline data used to estimate the behavior policy is suboptimal (Kumar 2019, Levine et al. 2020, Singh et al. 2023). Given that the collected human response data is not perfect as indicated in Section 4, the estimated behavior policy $\pi_{\text{ref}}$ can be suboptimal. An overly conservative distribution constraint between $\pi_\theta$ and $\pi_{\text{ref}}$, like the KL-divergence constraint in Equation 2, can render the learned policy $\pi_\theta$ to be suboptimal.

Given the KL-constrained objective in Equation 2, a closed-form solution of the optimal policy of Equation 1 can be derived using techniques similar to those in AWR (Peng et al. 2019) as follows:

$$\pi^*(y|x) = \frac{1}{Z(x)} \pi_{\text{ref}}(y|x) \exp\left(\frac{1}{\beta} r(x, y)\right), \tag{3}$$

where $Z(x) = \sum_y \pi_{\text{ref}}(y|x) \exp\left(\frac{1}{\beta} r(x, y)\right)$ represents the partition function. By further leveraging the analytical mapping between $\pi^*(y|x)$ and $r(x, y)$, the Bradley-Terry model (Bradley and Terry 1952) for human preference or reward modeling to train the reward function $r$ using response pair $(y_w, y_l)$, the preference policy optimization objective in Equation 1 can be derived (Rafailov et al. 2023). To summarize, the above analysis highlights that the excessive conservatism in the derived closed-form policy stems from the distributional policy constraint utilized in the offline RL objective (i.e., Equation 2).

**3.4.2 Relaxing Conservatism with Density Estimation-based Support Constraint**

Compared to distributional constraints, support constraints are less restrictive as they only necessitate that the learned policy assigns non-zero probability mass to actions with non-negligible behavior policy density (Kumar 2019, Levine et al. 2020). A more detailed survey of support constraint method is provided in Online Appendix A-26. Unlike the KL divergence constraint in Equation 2, support constraints do not demand a close density value matching between the learned policy and the behavior policy (Kumar et al. 2019). Thus, we propose to relax the distributional constraint in Equation 2 to a less restrictive support constraint, enabling better policy learning from suboptimal offline data. Specifically, we first estimate the density-based support of the offline data, denoted as $p_\psi(y|x,c)$, where $y$ represents the human response for the customer review in the offline data, and subsequently introduce a conservatism-relaxing term $\mathcal{J}_{cr}$ to update Equation 1 as follows:

$$\max_{\pi_\theta} \mathcal{J}_{pl} + \lambda \mathcal{J}_{cr}, \quad (4)$$

$$\mathcal{J}_{cr} = \mathbb{E}_{(x,c)\sim \mathcal{D}, y\sim \pi_\theta(\cdot|x,c)}[\log p_\psi(y|x,c)], \quad (5)$$

where $\lambda$ is the weight of the conservatism relaxing term. In Equation 4, while the learned policy is optimized under an initially conservative constraint, the second term $\mathcal{J}_{cr}$ encourages optimizing policies within the broader area identified by the support constraint, which extends beyond the restricted area considered by the original distributional policy constraint. Furthermore, our proposed conservatism relaxing term is designed to be plug-and-play, allowing for seamless integration into various preference optimization methods.

To estimate the density-based support of the offline data $p_\psi(y|x,c)$, we employ a variational autoencoder (VAE) (Kingma and Welling 2014, Sohn et al. 2015), which is one of the widely used density estimation methods (Fujimoto et al. 2019, Zhou et al. 2021). Specifically, we leverage the conditional variational autoencoder (CVAE) (Sohn et al. 2015) to estimate the density $p_\psi(y|x,c)$ by maximizing the evidence lower bound as follows:

$$\log p_\psi(y|x,c) \geq \mathbb{E}_{q_\phi(z|x,c,y)}\left[\log \frac{p_\psi(y,z|x,c)}{q_\phi(z|x,c,y)}\right], \quad (6)$$

$$= \mathbb{E}_{q_\phi(z|x,c,y)}[\log p_\psi(y|x,c,z)] - \mathbb{D}_{\text{KL}}[q_\phi(z|x,c,y)||p_\psi(z|x,c)],$$

$$\stackrel{\text{def}}{=} \mathcal{J}_{rec} - \mathcal{J}_{kl},$$

$$\stackrel{\text{def}}{=} \mathcal{L}_{ELBO}, \tag{7}$$

where $z$ is the latent variable; $\mathcal{J}_{rec}$ is the reconstruction term to encourage the model to reconstruct the response $y$; $\mathcal{J}_{KL}$ regularizes the posterior distribution $q_\phi(z|x,c,y)$ to be close to the prior distribution $p_\psi(z|x,c)$. Given the response $y$ contains a sequence of tokens $y = (y_1, \cdots, y_{|y|})$, we further decompose Equation 7 into timestep-wise evidence lower bound with the following assumptions about the generative and inference distribution in the CVAE model:

$$p_\psi(y,z|x,c) = \prod_{t=1}^{|y|} p_\psi(y_t|z_t, y_{<t}, x, c) p_\psi(z_t| y_{<t}, x, c), \tag{8}$$

$$q_\phi(z|x,c,y) = \prod_{t=1}^{|y|} q_\phi(z_t|y, x, c), \tag{9}$$

where $z = (z_1, \cdots, z_{|y|})$ is the sequence of latent variables and $y_{<t} = (y_1, \cdots, y_{t-1})$. Equation 8 and 9 follow a similar factorization of generative and inference distribution for sequential data modeling (Chung et al. 2016, Tang and Matteson 2021). As we employ the Transformer model (Vaswani et al. 2017) for the approximation of the generative and inference distribution (detailed in the model introduction in Online Appendix A-11), we do not explicitly model the dependency of latent variable $z_t$ and preceding latent variables. This approach preserves Transformer's inherent parallel computing capabilities, thereby maximizing computational efficiency in processing sequences. By integrating Equation 8 and 9 into Equation 6, we derive the timestep-wise evidence lower bound:

$$\mathcal{L}_{ELBO}(x,c,y) = \sum_{t=1}^{|y|} \mathbb{E}_{q_\phi(z_t|y,x,c)}[\log p_\psi(y_t|y_{<t}, z_t, x, c)] - \mathbb{D}_{KL}[q_\phi(z_t|y, x, c)||p_\psi(z_t|y_{<t}, x, c)]. \tag{10}$$

Given $\mathcal{L}_{ELBO}$ in Equation 10, it serves as the estimation of the density-based support of offline data $\log p_\psi(y|x,c)$ in Equation 5. We introduce the details of the CVAE model with transformer architecture (denoted as trans-CVAE) in Online Appendix A-11.

### 3.4.3 Theoretical Proof of Our Proposed Method for Relaxing Conservatism

In this section, we present a mathematically proved theorem to validate the superior theoretical guarantee of our proposed method for relaxing conservatism in offline preference learning. The theorem is stated as follows, with a detailed description and proof provided in the Online Appendix A-12.

**Theorem 3.4.3** Under the policy realizability and bounded reward assumption, with probability at least $1 - \delta$, our learned objective in Equation 4 produces a policy $\hat{\pi}$ such that for all policies $\pi^*$ simultaneously, we have the following bound of performance gap:

$$\mathcal{V}(\pi^*) - \mathcal{V}(\hat{\pi}) \leq (1+e^R)^2 \sqrt{\frac{2C^{\pi^*}\log(|\Pi|/\delta)}{n}} + (\beta + \lambda)C^{\pi^*} + \frac{(1+e^R)^4 \log(|\Pi|/\delta)}{2n\lambda},$$

where the performance metric $\mathcal{V}(\pi) \stackrel{\text{def}}{=} \mathbb{E}_{(x,c)\sim\rho, y\sim\pi(\cdot|x)} r^*(x,c,y)$ is the expected true reward $r^*(x,c,y)$ over the review and context distribution $\rho$ and model generated responses. $R$ is the maximum reward value; $|\Pi|$ is the size of policy class; $n$ is the preference finetuning data size. $C^{\pi} \stackrel{\text{def}}{=} \mathbb{E}_\pi \left[\frac{1}{\pi_{\text{ref}}}\right]$ is the coverage coefficient for policy $\pi$ and $C^{\pi}$ quantifies the extent to which $\pi$ is covered by $\pi_{\text{ref}}$ or the offline data distribution (Gabbianelli et al. 2024).

*Remarks*: Theorem 3.4.3 informs that our proposed method achieves a performance guarantee based on favorable single-policy coverage $C^{\pi^*}$ when compared with the optimal policy $\pi^*$, while the performance guarantee of DPO is based on undesirable all-policy coverage $\max_\pi C^\pi$ (detailed in Theorem A-12-1 in Online Appendix A-12). The literature on offline RL theory informs that the performance bound based on single-policy coverage is significantly better than the all-policy coverage (L. Wang et al. 2024). Specifically, if the optimal policy $\pi^*$ has a good coverage under the offline data, the performance gap of our method would be small. However, the theoretical performance gap of DPO will be still large based on all-policy coverage due to the inclusion of low-quality policy with poor coverage. Therefore, our proposed method for offline preference learning offers a superior theoretical guarantee compared to DPO.

Due to space limit, we include the detailed LLM finetuning algorithm in Online Appendix A-14.

## 4 EMPIRICAL EVALUATION

We evaluate our proposed LLM fine-tuning method against several state-of-the-art baseline methods on the widely recognized hotel review response task (Kew and Volk 2022, Ravichandran and Deng 2023, Deng and Ravichandran 2023), where previous research has demonstrated a significant impact of online review responses on hotel financial performance (Xie et al. 2017, Kumar et al. 2018).

### 4.1 Dataset

We utilize a public dataset of hotel review-response pairs collected from TripAdvisor, a leading travel review platform (Kew and Volk 2022). The original dataset contains 500,000 unique review-response pairs, where all responses are written by humans. Given that customer review ratings range from 1 to 5, we classify reviews with ratings below 3 as negative, and those with ratings of higher than 3 as positive (Ravichandran and Deng 2023, Deng and Ravichandran 2023). Recent studies have emphasized that the quality of fine-tuning data is more crucial than quantity, exemplified by the 1,000 manually curated high-quality instructional data used by LIMA (Zhou et al. 2023). Thus, we perform the following data curation to create training sets from the original dataset. We particularly focus on curating data for negative review-response pairs, as responding to negative reviews poses greater challenges than positive. Our data curation process involves multiple steps. Firstly, acknowledging that longer responses generally provide more comprehensive redressal of customers' complaints, we sort the negative review responses by length in descending order, capping the maximum length at 400 words to avoid excessively long responses. Second, we filter out those human responses of low quality, including those that dismiss consumer complaints or even include personal attacks on consumers. Inspired by recent LLM-based instruction fine-tuning data selection methods (L. Chen et al. 2023), we employ LLaMA2-70B-Chat (Touvron et al., 2023) to assess the quality of human responses based on their helpfulness, relevance, accuracy, and level of detail, using a scoring scale from 0 to 5. Responses scored below 3 are filtered out. Finally, we select the top 1,000 negative review-response pairs from the remaining sorted list and randomly sample 200 positive review-response pairs, both of which form the initial training dataset; we randomly sample 100 negative review-response pairs and 20 positive pairs as a validation set for model selection; for the test set, we randomly select 2,000 negative and 1,000 positive pairs that have not appeared in the training and validation sets. In addition, we randomly select 50 negative pairs for each review type from the test set (negative review type classification is detailed in Table 2), resulting in a total of 200 negative review-response pairs, for human evaluation; considering that most positive reviews pertain to the first category of positive review types and that responses to positive reviews are generally less challenging, we select 50 positive pairs from the test set for human evaluation.

### 4.2 Benchmark Methods

Based on a comprehensive review of the literature, we have selected seven state-of-the-art baseline methods for comparison. Given that our fine-tuning approach utilizes *LLaMA2-70B-Chat* [1](Touvron et al., 2023), we first benchmark the performance of our method against this prominent open-source LLM. Second, we compare our approach with a recently proposed iterative prompt engineering method (abbreviated as *AutoPrompt*), which employs a self-refine technique (Madaan et al. 2023) to optimize the prompt for generating review responses (Azov et al. 2024). We apply this method to LLaMA2-70B-Chat for prompt optimization. Third, we adapt a recently proposed Supervised Fine-Tuning method (Zhou et al. 2023), abbreviated as *SFT*, to fine-tune LLaMA2-70B-Chat with our dataset. Fourth, we compare our method with a prompt-tuning (Liu et al., 2023) method. Prompt-tuning (Liu et al., 2023) adds trainable prompt embeddings to the input that is optimized by a prompt encoder to find a better prompt. We use prompt-tuning to fine-tune LLaMA2-70B-Chat with the supervised learning loss on our dataset. Fifth, we compare our method with the retrieval augmented generation (RAG) on LLaMA2-70B-Chat using our extracted contextual information to ensure a fair comparison. Specifically, we first aggregate the extracted context information in the training data to the hotel level . Then, we apply RAG (Gao et al. 2024) on the context information for each test customer review to generate the managerial response. For those hotels in the test data without any context information, we directly prompt LLM to generate the response. Considering GPT-4 (OpenAI et al. 2024) as the most advanced commercial LLM currently available, which is a closed source LLM and does not allow finetuning with our preference pairs data, we compare our method against *GPT-4*. Additionally, we conduct preference finetuning for the original *DPO* method using our designed preference pairs, which yields the last baseline method. These selected baselines represent the state of the art in the broad area of prompt engineering and finetuning-based methods, encompassing both open-source and closed-source advanced LLMs. Our method is abbreviated as *TPO-CAR*. A summary of these methods is provided in Table 4.

---

[1] The fine-tuning of LLaMA2-70B-Chat using our method was completed prior to the release of LLaMA3. Given the economic costs associated with renting GPUs, we did not apply our fine-tuning method to LLaMA3, although we anticipate improved performance with this newer model. It is important to note that our finetuning method is LLM-agnostic and can be applied to any available LLMs.

Due to space limit, we include the implementation details of all competing methods in Online Appendix A-15. For a fair comparison, we use the same prompt instruction and concatenate this prompt instruction with the customer review for baseline methods SFT, LLaMA2-70B-Chat, GPT-4 and our method TPO-CAR.

Table 4. A summary of competing methods

| Method | Abbreviation | Category |
|---|---|---|
| LLaMA2-70B-Chat (Touvron et al. 2023) | LLaMA2-70B-Chat | Preference finetuning |
| Iterative prompt engineering on LLaMA2-70B-Chat using self-refinement for online review response generation (Azov et al. 2024) | AutoPrompt | Prompt engineering |
| Supervised finetuning method adapted for online review response generation to fine-tune LLaMA2-70B-Chat (Zhou et al. 2023) | SFT | Supervised finetuning |
| GPT-4 (OpenAI et al. 2024) | GPT-4 | Preference finetuning |
| Adapted DPO using our constructed preference pairs to fine-tune LLaMA2-70B-Chat (Rafailov et al. 2023) | DPO | Preference finetuning |
| Prompt tuning to fine-tune LLaMA2-70B-Chat for online review response generation (X. Liu et al. 2023) | Prompt-tuning | Prompt tuning |
| Adapted retrieval-augmented generation (RAG) using our extracted context information from training data (Gao et al. 2024) on LLaMA2-70B-Chat | RAG | Retrieval-augmented prompting |
| Context-augmented theory-driven preference optimization with relaxed conservatism to fine-tune LLaMA2-70B-Chat (**our method**) | TPO-CAR | Preference finetuning |

### 4.3 Evaluation Using Objective Evaluation Metric

We employ objective evaluation metrics to assess the quality of responses generated by different methods and present multiple qualitative analyses for interpreting the effectiveness of our designed modules.

**4.3.1 Objective Evaluation Metric**

We utilize BERTScore (Zhang et al. 2020), a widely used automatic evaluation metric for evaluating generated responses (Van Veen et al. 2024). Traditional automatic evaluation metrics such as BLEU or ROUGE (Zhang et al. 2020) only rely on the surface form similarity by calculating the n-gram overlap between the model's generations and human references. These n-gram match metrics, however, often fail to match paraphrases robustly and do not adequately capture distant dependencies or account for semantically critical ordering changes (Zhang et al. 2020). In contrast, BERTScore leverages pretrained contextual embeddings to evaluate the semantic similarity between a model-generated response and a human reference. This is achieved by computing the sum of cosine similarities between the embeddings of corresponding tokens, a method that has demonstrated high correlation with human judgment (Zhang et al. 2020). Besides, recent studies have investigated the use of GPT-4 as a judge to evaluate responses generated by different models (Wang et al. 2023). However, such LLM-based evaluators have exhibited biases towards their own outputs and the order in which responses are presented (Panickssery et al. 2024). Considering GPT-4's role as one of the competing methods and the sensitivity of LLMs to textual evaluation instructions and inputs (Bowman 2023), we have opted not to use GPT-4-based evaluator as our objective metric. Given a tokenized model-generated response $\hat{y} = \{\hat{y}_1, \cdots, \hat{y}_{|\hat{y}|}\}$ and the tokenized human response $y = \{y_1, \cdots, y_{|y|}\}$ for the same customer review, BERTScore of the recall, precision, and F1 scores are calculated as: $R_{\text{bert}} = \frac{1}{|y|}\sum_{y_i \in y} \max_{\hat{y}_j \in \hat{y}} \mathbf{y}_i^\top \hat{\mathbf{y}}_j$, $P_{\text{bert}} = \frac{1}{|\hat{y}|}\sum_{\hat{y}_j \in \hat{y}} \max_{y_i \in y} \mathbf{y}_i^\top \hat{\mathbf{y}}_j$, $F_{\text{bert}} = 2\frac{P_{\text{bert}} \times R_{\text{bert}}}{P_{\text{bert}} + R_{\text{bert}}}$, where $\mathbf{y}_i$ and $\hat{\mathbf{y}}_j$ are the BERT embedding for the $i$-th token of $\hat{y}$ and $j$-th token of $y$, respectively. We further use baseline rescaling to enhance the readability of the BERTScore, as suggested in Zhang et al. (2020). All these three metrics measure the quality of model-generated responses via comparing with human responses, and a higher value indicates a better performance.

### 4.3.2 Evaluation Results

We report the evaluation results of compared methods for different negative and positive review types in Table 5 and 6, respectively. Additionally, we provide "overall" results encompassing all review categories. We have the following observations from the results. *First*, our method significantly outperforms all seven baseline methods, including the advanced GPT-4 model, across different review types and objective evaluation metrics. Specifically, considering the "overall" results, our method significantly outperforms the best baseline method by 11.2% on $R_{\text{bert}}$, 21.4% on $P_{\text{bert}}$ and 17.0% on $F_{\text{bert}}$, achieving these gains with a statistical significance level of $p<0.01$. Notably, AutoPrompt emerges as the superior baseline method according to $R_{\text{bert}}$ metric, while GPT-4 excels in $P_{\text{bert}}$ and $F_{\text{bert}}$ metrics. *Second*, we observe that SFT performs notably worse on the negative review response generation. A primary contributor to this issue is the presence of factual elements in review responses that do not appear in customer's original reviews. This discrepancy in the training dataset leads the model to learn to hallucinate the unknown facts beyond the given customer review. Similar reason applies to the poor performance of the prompt-tuning method with supervised learning loss. Conversely, for positive reviews, where the responses in the training data are less fact-intensive, the SFT method performs considerably better, as shown in Table 6. The inferior performance of SFT method reflects the non-trivial challenge of LLM finetuning for real-world business applications and underscores the effectiveness of our novel LLM fine-tuning approach. *Third*, while the AutoPrompt method, which employs automatic prompt engineering with self-refinement, improves performance in generating responses to negative reviews compared to the base LLaMA2-70B-Chat model, it still falls significantly short of our method (i.e., TPO-CAR outperforms it by 30.4% on the $F_{bert}$ metric). This indicates the limitations of prompt engineering compared to our advanced fine-tuning strategy. Besides, we observe AutoPrompt underperforms the base model in positive review response evaluations, as the optimized prompt template generated by AutoPrompt mainly contains the response strategies for the customer concerns in negative reviews. *Fourth*, preference finetuning using DPO with our designed preference pairs outperforms the AutoPrompt method and the RAG method. This comparison highlights the superiority of preference fine-tuning over prompt engineering methods for review response generation. Furthermore, our proposed method significantly surpasses the DPO method by 26.8% on the $F_{bert}$ metric in the "Overall" evaluations of negative reviews. This result underscores the enhanced efficacy of our curriculum learning approach and the relaxation of conservatism in offline preference learning. *In summary*, the superior performance of our method, TPO-CAR, validates the novel design of context augmentation, theory-driven preference optimization (preference pair construction, curriculum learning), and relaxed conservatism in offline preference optimization.

To further demonstrate the generalizability of our method, we use our proposed method to fine-tune a more recent LLaMA3.1-70B-Chat model and include the results in Online Appendix A-27.

Table 5. Evaluation results on different negative review types ("overall" includes all the review types). The best-performing model on each metric is highlighted in bold, % indicates improvement over the best baselines (underlined), *$p<0.1$, **$p<0.05$, ***$p<0.01$ denotes the statistical significance of the difference between our TPO-CAR method and the best baseline method.

| Review type | Methods | $R_{bert}$ | $P_{bert}$ | $F_{bert}$ |
|---|---|---|---|---|
| | **TPO-CAR** | **0.1614**<br>+6.0% | **0.2352**<br>+6.5% | **0.1988**<br>+6.0% |
| Negative Review Type 1 | SFT | 0.1428 | 0.0631 | 0.1037 |
| | AutoPrompt | 0.1545 | 0.1600 | 0.1582 |
| | Prompt-tuning | 0.1433 | 0.1561 | 0.1505 |
| | RAG | 0.1490 | 0.1757 | 0.1634 |
| | DPO | 0.1469 | 0.1847 | 0.1666 |
| | LLaMA2-70B-Chat | 0.1297 | 0.1669 | 0.1492 |
| | GPT-4 | 0.1523 | 0.2209 | 0.1875 |
| | **TPO-CAR** | **0.1526**<br>+17.4% | **0.2560**<br>+25.2% | **0.2047**<br>+21.8% |

| Review type | Methods | | | |
|---|---|---|---|---|
| Negative Review Type 2 | SFT | 0.1400 | 0.0842 | 0.1131 |
| | AutoPrompt | 0.1377 | 0.1763 | 0.1581 |
| | Prompt-tuning | 0.1193 | 0.1640 | 0.1425 |
| | RAG | 0.1286 | 0.1759 | 0.1533 |
| | DPO | 0.1319 | 0.1864 | 0.1601 |
| | LLaMA2-70B-Chat | 0.1097 | 0.1813 | 0.1464 |
| | GPT-4 | 0.1300 | <u>0.2045</u> | <u>0.1682</u> |
| Negative Review Type 3 | **TPO-CAR** | **0.1644**\*\*<br>**+6.9%** | **0.2315**\*\*<br>**+7.2%** | **0.1985**\*\*\*<br>**+6.8%** |
| | SFT | 0.1450 | 0.0650 | 0.1057 |
| | AutoPrompt | <u>0.1576</u> | 0.1673 | 0.1634 |
| | Prompt-tuning | 0.1416 | 0.1600 | 0.1517 |
| | RAG | 0.1466 | 0.1769 | 0.1628 |
| | DPO | 0.1451 | 0.1805 | 0.1637 |
| | LLaMA2-70B-Chat | 0.1400 | 0.1846 | 0.1631 |
| | GPT-4 | 0.1538 | <u>0.2159</u> | <u>0.1857</u> |
| Negative Review Type 4 | **TPO-CAR** | **0.1540**\*\*\*<br>**+9.3%** | **0.2825**\*\*\*<br>**+26.8%** | **0.2182**\*\*\*<br>**+19.5%** |
| | SFT | 0.1356 | 0.0787 | 0.1081 |
| | AutoPrompt | <u>0.1433</u> | 0.1777 | 0.1616 |
| | Prompt-tuning | 0.1233 | 0.1753 | 0.1500 |
| | RAG | 0.1378 | 0.1838 | 0.1618 |
| | DPO | 0.1368 | 0.1979 | 0.1681 |
| | LLaMA2-70B-Chat | 0.1211 | 0.1957 | 0.1592 |
| | GPT-4 | 0.1409 | <u>0.2228</u> | <u>0.1827</u> |
| Overall | **TPO-CAR** | **0.1556**\*\*\*<br>**+11.2%** | **0.2613**\*\*\*<br>**+21.4%** | **0.2087**\*\*\*<br>**+17.0%** |
| | SFT | 0.1391 | 0.0771 | 0.1091 |
| | AutoPrompt | <u>0.1443</u> | 0.1738 | 0.1601 |
| | Prompt-tuning | 0.1265 | 0.1671 | 0.1476 |
| | RAG | 0.1369 | 0.1792 | 0.1591 |
| | DPO | 0.1372 | 0.1902 | 0.1646 |
| | LLaMA2-70B-Chat | 0.1203 | 0.1857 | 0.1538 |
| | GPT-4 | 0.1399 | <u>0.2153</u> | <u>0.1784</u> |

Table 6. Evaluation results on different positive review types (positive review test data only contains positive review types 1 and 2, "overall" includes all the review types). The best-performing model on each metric is highlighted in bold, % indicates improvement over the best baselines (underlined), $*p<0.1, **p<0.05, ***p<0.01$ denotes the statistical significance of the difference between our TPO-CAR method and the best baseline method.

| Review type | Methods | $R_{bert}$ | $P_{bert}$ | $F_{bert}$ |
|---|---|---|---|---|
| Positive Review Type 1 | **TPO-CAR** | **0.3192**<br>**+0.5%** | **0.4537**\*\*\*<br>**+27.9%** | **0.3856**\*\*\*<br>**+35.9%** |
| | SFT | 0.2655 | 0.2986 | 0.2819 |
| | AutoPrompt | 0.3088 | 0.0661 | 0.1852 |
| | Prompt-tuning | 0.2264 | 0.1808 | 0.2033 |
| | RAG | 0.3055 | 0.1158 | 0.2095 |
| | DPO | 0.2103 | <u>0.3547</u> | 0.2816 |
| | LLaMA2-70B-Chat | 0.3109 | 0.1625 | 0.2362 |
| | GPT-4 | <u>0.3176</u> | 0.2494 | <u>0.2837</u> |
| Positive Review Type 2 | **TPO-CAR** | **0.3605**<br>**+4.3%** | **0.4665**\*\*\*<br>**+18.4%** | **0.4135**\*\*\*<br>**+27.1%** |
| | SFT | 0.3158 | 0.3330 | 0.3246 |
| | AutoPrompt | 0.3124 | 0.0263 | 0.1657 |
| | Prompt-tuning | 0.2367 | 0.1644 | 0.1980 |

|  | RAG | 0.3218 | 0.1130 | 0.2158 |
|  | DPO | 0.2336 | <u>0.3941</u> | 0.3128 |
|  | LLaMA2-70B-Chat | 0.3266 | 0.1627 | 0.2440 |
|  | GPT-4 | <u>0.3456</u> | 0.3041 | <u>0.3253</u> |
|  | **TPO-CAR** | **0.3205** <br> **+0.6%** | **0.4540\*\*\*** <br> **+27.6%** | **0.3864\*\*\*** <br> **+35.6%** |
| Overall | SFT | 0.2671 | 0.2997 | 0.2832 |
|  | AutoPrompt | 0.3089 | 0.0648 | 0.1846 |
|  | Prompt-tuning | 0.2267 | 0.1803 | 0.2031 |
|  | RAG | 0.3060 | 0.1157 | 0.2097 |
|  | DPO | 0.2111 | <u>0.3559</u> | 0.2825 |
|  | LLaMA2-70B-Chat | 0.3114 | 0.1625 | 0.2365 |
|  | GPT-4 | <u>0.3185</u> | 0.2551 | <u>0.2850</u> |

#### 4.3.3 Ablation Study

To elucidate the individual contributions of each designed component within our proposed TPO-CAR method, we conduct an ablation study by removing each of the four key modules: context augmentation, preference pair construction, curriculum learning, and relaxing conservatism. Additionally, we evaluate the statistical significance of the performance discrepancies between our full TPO-CAR model and each variant. The results (shown in Table 7 and 8) yield several critical insights. *First*, the exclusion of any module results in substantial performance declines across all three objective evaluation metrics, spanning both negative and positive review types. Specifically, for the overall negative review response assessed by the $F_{\text{bert}}$ metric, our full model TPO-CAR significantly outperforms the model variant w/o context augmentation by 68.9%, w/o preference pair construction by 21.6%, w/o curriculum learning by 12.2% and w/o relaxing conservatism by 17.7%. *Second*, among the model variants, those lacking context augmentation or preference pair construction exhibits the most pronounced decreases in overall performance for negative and positive review response generation, respectively. This is primarily because negative responses in the training data contain a greater prevalence of factual content. Removing the context augmentation module will significantly introduce the hallucination in the negative review response generations, thereby detrimentally affecting model performance. *Third*, the model variant w/o preference pair construction demonstrates the second most significant performance decline for negative review response generation and the largest for positive review response generation. For instance, our full model significantly outperforms this model variant by 27.9% on the $F_{\text{bert}}$ metric for negative review type 2, underscoring the effectiveness of incorporating preference pair construction to enhance emotional over rational cues in response generation, aligning with human preferences. In summary, this ablation study clearly demonstrates the effectiveness of each designed module in our TPO-CAR method.

Table 7. Ablation study on different negative review types ("overall" includes all the review types). The best-performing model on each metric is highlighted in bold, *p<0.1,**p<0.05,***p<0.01 denotes the statistical significance of the difference between our TPO-CAR model and each model variant.

| Review type | Methods | $R_{bert}$ | $P_{bert}$ | $F_{bert}$ |
|---|---|---|---|---|
| Negative Review Type 1 | TPO-CAR | **0.1614** | **0.2352** | **0.1988** |
| | w/o context augmentation | 0.1545** | 0.0826*** | 0.1194*** |
| | w/o preference pair construction | 0.1316*** | 0.2041*** | 0.1684*** |
| | w/o curriculum learning | 0.1362*** | 0.2245** | 0.1809*** |
| | w/o relaxing conservatism | 0.1386*** | 0.2147*** | 0.1772*** |
| Negative Review Type 2 | TPO-CAR | **0.1526** | **0.2560** | **0.2047** |
| | w/o context augmentation | 0.1470*** | 0.1122*** | 0.1307*** |
| | w/o preference pair construction | 0.1097*** | 0.2096*** | 0.1601*** |
| | w/o curriculum learning | 0.1194*** | 0.2225*** | 0.1714*** |
| | w/o relaxing conservatism | 0.1194*** | 0.2214*** | 0.1709*** |
| Negative Review Type 3 | TPO-CAR | **0.1644** | **0.2315** | **0.1985** |
| | w/o context augmentation | 0.1544*** | 0.0943*** | 0.1253*** |
| | w/o preference pair construction | 0.1287*** | 0.2057*** | 0.1677*** |
| | w/o curriculum learning | 0.1353*** | 0.2184** | 0.1774*** |
| | w/o relaxing conservatism | 0.1334*** | 0.2164*** | 0.1754*** |
| Negative Review Type 4 | TPO-CAR | **0.1540** | **0.2825** | **0.2182** |
| | w/o context augmentation | 0.1441*** | 0.0904*** | 0.1182*** |
| | w/o preference pair construction | 0.1223*** | 0.2455*** | 0.1840*** |
| | w/o curriculum learning | 0.1376*** | 0.2681*** | 0.2029*** |
| | w/o relaxing conservatism | 0.1231*** | 0.2435*** | 0.1834*** |
| Overall | TPO-CAR | **0.1556** | **0.2613** | **0.2087** |
| | w/o context augmentation | 0.1476*** | 0.0976*** | 0.1236*** |
| | w/o preference pair construction | 0.1197*** | 0.2230*** | 0.1717*** |
| | w/o curriculum learning | 0.1307*** | 0.2408*** | 0.1860*** |
| | w/o relaxing conservatism | 0.1249*** | 0.2289*** | 0.1773*** |

Table 8. Ablation study on different positive review types (positive review test data only contains positive review types 1 and 2, "overall" includes all the review types). The best-performing model on each metric is highlighted in bold, *p<0.1,**p<0.05,***p<0.01 denotes the statistical significance of the difference between our TPO-CAR model and each model variant.

| Review type | Methods | $R_{bert}$ | $P_{bert}$ | $F_{bert}$ |
|---|---|---|---|---|
| Positive Review Type 1 | TPO-CAR | **0.3192** | **0.4537** | **0.3856** |
| | w/o context augmentation | 0.3105*** | 0.3949*** | 0.3522*** |
| | w/o preference pair construction | 0.2483*** | 0.4251*** | 0.3353*** |
| | w/o curriculum learning | 0.2520*** | 0.4372*** | 0.3431*** |
| | w/o relaxing conservatism | 0.2534*** | 0.4344*** | 0.3424*** |
| | TPO-CAR | **0.3605** | **0.4665** | **0.4135** |
| | w/o context augmentation | 0.3427 | 0.4111*** | 0.3773** |
| | w/o preference pair construction | 0.2786*** | 0.4432 | 0.3601*** |
| Positive Review Type 2 | w/o curriculum learning | 0.2964*** | 0.4623 | 0.3784*** |
|  | w/o relaxing conservatism | 0.3005*** | 0.4496 | 0.3744*** |
| Overall | TPO-CAR | **0.3205** | **0.4540** | **0.3864** |
|  | w/o context augmentation | 0.3115*** | 0.3954*** | 0.3530*** |
|  | w/o preference pair construction | 0.2493*** | 0.4257*** | 0.3360*** |
|  | w/o curriculum learning | 0.2534*** | 0.4380*** | 0.3442*** |
|  | w/o relaxing conservatism | 0.2549*** | 0.4349*** | 0.3434*** |

**4.3.4 Qualitative Analysis of the Theory-driven Preference Optimization**

In this section, we present a qualitative analysis to interpret the effectiveness of the theory-driven preference optimization in our proposed method. This optimization is grounded on the construction of preference pairs,

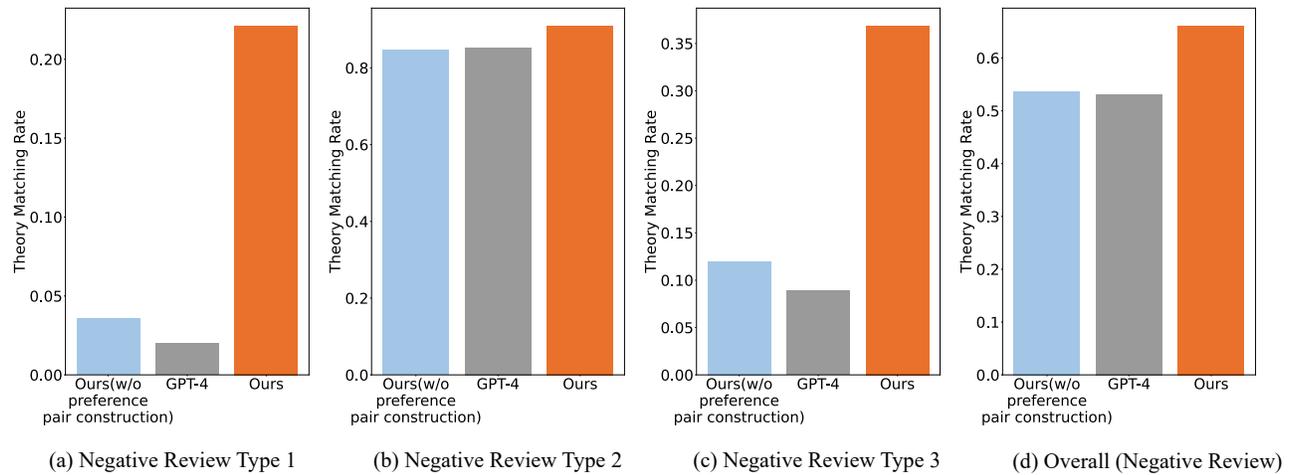

(a) Negative Review Type 1  (b) Negative Review Type 2  (c) Negative Review Type 3  (d) Overall (Negative Review)

Figure 3. Theory matching rate of generated responses for negative reviews by different methods. Responses generated by TPO-CAR (our full model) match the theory-suggested responses much better than its variant and GPT-4.

which are built upon established theoretical frameworks concerning optimal managerial response generation. We calculate *the theory matching rate* as the proportion of responses that align with the theoretically advised response type for each review category, as detailed in Sections 3.3.1 and 3.3.2. For instance, regarding negative review type 1, characterized by greater procedural than interactional unfairness, the theoretical framework advocates for responses that prioritize rational over emotional cues. The detailed method for identifying the rational and emotional cues in the responses is shown in Online Appendix A-16. Due to space limit, we present theory matching rates for our model, the model variant without preference pair construction, and the GPT-4 baseline across various negative review types in Figure 3. Theory matching rate for positive reviews is shown in Online Appendix A-17. As shown in Figure 3, our full model achieves a significantly higher theory matching rate across different negative review types than our model variant and the GPT-4 baseline. Specifically, for negative review type 1, the theory matching rates for the model variant (without preference pair construction) and GPT-4 are notably low, at 0.036 and 0.020 respectively. In contrast, our full model exhibits a theory matching rate of 0.221, significantly enhancing the proportion of responses that align with the theoretical preference for more rational cues over emotional cues. *In summary*, this analysis of theory matching rates effectively highlights the efficacy of our proposed theory-driven preference optimization, which is meticulously designed based on theory-guided preference pair construction. We further provide detailed response generation examples in the Online Appendix A-18.

### 4.3.5 Qualitative Analysis of Context Augmentation to Mitigate LLM Hallucination

Table 9. Average number of hallucinated facts across human annotators in generated responses over different negative review types.

| Methods | Average number of hallucinated facts | | | | |
| --- | --- | --- | --- | --- | --- |
| | Negative Review Type 1 | Negative Review Type 2 | Negative Review Type 3 | Negative Review Type 4 | Overall |
| TPO-CAR | 0.200 | 0.000 | 0.000 | 0.067 | 0.067 |
| TPO-CAR (w/o context augmentation) | 6.200 | 5.767 | 5.233 | 6.100 | 5.825 |
| RAG | 0.300 | 0.433 | 0.200 | 0.200 | 0.283 |

In this section, we conduct a qualitative analysis to interpret how the designed context augmentation module mitigates hallucination in LLM responses. Given the challenge of detecting hallucinations within review responses, we engage three human annotators with sufficient hotel accommodation experiences to identify objective facts within the responses that either do not appear in the customer review or cannot be logically inferred from it. To aid in this task, annotators can utilize the search tool to facilitate the identification of hallucinated facts. Consequently, for each test review and its associated generated response, the annotator compiles a list of hallucinated facts. Owing to the substantial costs associated with human labeling, our analysis focuses on negative review response generation. We randomly sample 10 reviews for each category of negative review from the test data and let the annotator label them. Our findings (shown in Table 9) indicate that the inclusion of the context augmentation module significantly diminishes the occurrence of hallucinated facts in responses across these review types. Specifically, the average number of hallucinated facts per response decreased dramatically from 5.825 to 0.067 when utilizing our context augmentation module. An illustrative example comparing response generation by these two methods for a test review is depicted in Online Appendix A-19. Furthermore, our proposed method outperforms the RAG benchmark method in minimizing hallucinations in generated responses.

### 4.4 Human Evaluation

We further assess the responses generated by our method through human evaluations, which is done by comparing with the responses generated by the leading baseline method, GPT-4, and hotel managers.

#### 4.4.1 Human Evaluation Setting

The human evaluation metrics are designed based on the relevant theory and literature. For negative reviews, guided by the theoretical framework of justice theory for explaining customer complaint behaviors (Tax et al. 1998, Ravichandran and Deng 2023), we consider three important elements of justice theory: distributive unfairness, procedural unfairness and interactional unfairness as detailed in Section 3.3.1. Evaluators are tasked with assessing how effectively the responses to negative reviews address each of these dimensions of customer complaints. Furthermore, the overall response quality is included as an additional evaluation metric. For positive reviews, the primary metric is the overall response quality. The detailed definitions of each evaluation dimension are provided in the human evaluation interface (as shown in Online Appendix A-20).

We randomly select 50 negative reviews for each review type from the test set, resulting in a total of 200 reviews for human evaluations. Similarly, 50 positive reviews are randomly selected for human evaluations. Three human evaluators, all of whom have sufficient experience in hotel accommodations, are recruited for this task. Each evaluator is presented with two anonymized responses (either Ours versus GPT-4's, or Ours versus Human's) in a randomized order for each review—negative or positive. They are required to compare the quality of these responses along each of the specified evaluation dimensions. The human evaluation interface is depicted in Online Appendix A-20.

#### 4.4.2 Human Evaluation Results

After collecting the human evaluation data, we compute the win rate as the average proportion of evaluations where our method's response is favored across all reviewers and reviews. This approach was also applied to calculate the Tie and Loss rates for our model. Figure 4 displays the human evaluation results for the comparison of our method against both GPT-4 model and human-written responses for negative reviews. The results distinctly indicate that our method's win rates significantly surpass those of the GPT-4 model and human-written responses across various evaluation dimensions. For instance, the win rate on the "overall" dimension for our model is 71.8%, while it stands at 7.2% for the GPT-4 model. When comparing the responses by our model with human responses, the win rate of our model slightly decreases but remains significantly superior across different evaluation dimensions.

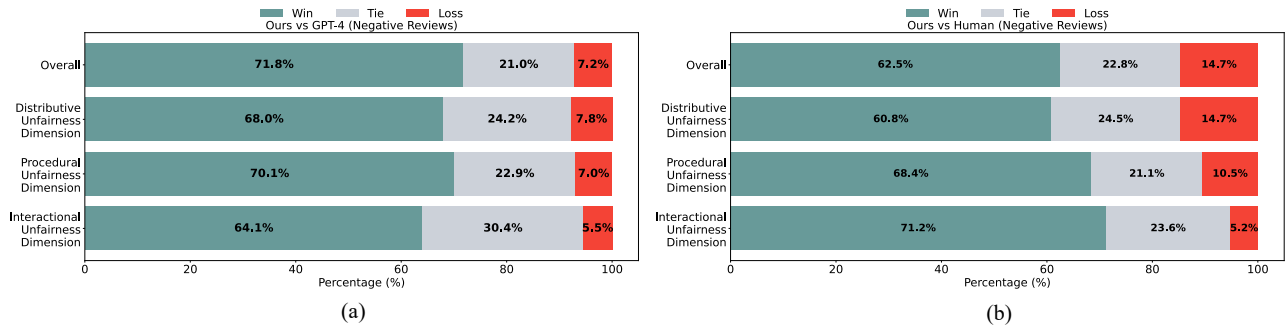

Figure 4. Human evaluation results across different evaluation dimensions for negative customer reviews: (a) Ours vs GPT-4, (b) Ours vs Human.

Furthermore, we assess the inter-rater reliability for each of the four evaluation dimensions in the two types of model comparisons using the widely recognized Cronbach's Alpha metric (Ravichandran and Deng 2023). All Cronbach's Alpha values exceed the acceptable threshold of 0.7, underscoring consistent evaluator agreement. The human evaluation results across different types of negative reviews are detailed in Online Appendix A-21, where we observe a consistent superiority of our proposed method. In Online Appendix A-25, we further test our model against the best baseline GPT-4 model using more recent review without human managerial responses. These human evaluation results unequivocally demonstrate the superiority of our method in addressing various dimensions of customer complaints compared with both GPT-4 and human-written responses, from a human perspective.

For positive reviews, the win, tie, and loss rates are 32.0%, 45.3%, and 22.7%, respectively, in the evaluation of our model versus GPT-4. In comparisons of our model against human responses, these rates are 50.0%, 11.3%, and 38.7%, respectively. These outcomes substantiate the superiority of our model in generating responses to positive reviews. When comparing these results with those on negative reviews, the relatively smaller win rate gap between our method and the compared models may be attributable to the inherently less challenging nature of generating responses to positive reviews.

In conclusion, the human evaluation results demonstrate that our proposed method can significantly enhance user satisfaction with the generated responses, subsequently yielding substantial business benefits. We further provide recommendations for the real-world deployment of our proposed method in Online Appendix A-24.

## 5    CONCLUSION AND DISCUSSION

### 5.1 Research Contribution

This paper addresses an important problem, finetuning LLM for customer review responses. With our designed LLM finetuning solution, we make the following contributions to the literature. (1) We proposed a novel LLM finetuning method to automatically generate customer review responses of good quality, which yields significant benefits to multiple stakeholders. Our study makes a meaningful contribution to the field of computational design science in Information Systems (Padmanabhan et al. 2022, Abbasi et al. 2024), which emphasizes the development of innovative computational algorithms and machine learning models to address challenges associated with information systems. (2) To tackle the hallucination challenge when finetuning LLM for business applications, we first identified the hallucination source and then proposed a novel context augmentation approach to bridge the context information gap in the finetuning data. This proposed approach is simple and effective in mitigating LLM's hallucinations. It expands the current research literature in LLM hallucination and the idea could be adapted to mitigate hallucinations in other domain-specific LLMs. (3) Most existing LLM preference finetuning methods (e.g., DPO) rely on preference pair data and ignore domain-specific knowledge. However, in online review domain, only a preferred response (i.e., human response) exists for a given customer review. To tackle this challenge, we proposed a novel theory-driven preference learning method that automatically constructs preference pairs and fuses with the DPO technique for LLM finetuning. This novel theory-driven preference learning seamlessly integrates the domain-specific knowledge into the LLM finetuning process to better align LLM with domain-specific human preferences.

Our proposed method exemplifies the novel application of IS theory for LLM finetuning. Our method also includes a curriculum learning design for more effectively fine-tuning the LLM. The curriculum learning design is applicable to finetuning LLMs for other applications. (4) We developed a novel density estimation-based support constraint approach to relax the conservatism in the DPO-based policy learning to avoid being overly restricted to the suboptimal offline finetuning data. Our theoretical analyses validate the superiority of our approach to DPO-based policy learning. This developed approach fundamentally expands the technical frontier in DPO-based LLM finetuning, making a significant contribution to the general LLM field. We conducted extensive experiments, which include evaluations based on objective metrics, human assessments, and qualitative analyses, to validate the superiority of our proposed LLM fine-tuning methodology. In addition, we provided a recommendation of two types of hybrid approaches for deploying our solution in the real-world systems (e.g., TripAdvisor).

## 5.2 Practical Implication

This study has practical implications and values for several stakeholders. *First*, our developed artifact could significantly benefit firms such as hotel companies by reducing costly human labor or time required to respond to customer reviews. Given the exponential growth in the volume of online reviews and the critical importance of timely responses to these reviews (Ravichandran and Deng 2023, Deng and Ravichandran 2023), our fine-tuned LLM, in conjunction with the recommended hybrid approaches for deployment, is poised to deliver considerable economic benefits to firms. It will also enhance their customer relationship management through addressing customer complaints more efficiently. *Second*, our solution, upon deployment, will help enhance customer satisfaction, particularly regarding customer complaints. Currently, due to the sheer volume of customer reviews, a significant proportion of negative reviews remain unaddressed (Howarth 2023), which adversely affects customer satisfaction. By deploying our fine-tuned LLM using our recommended hybrid approaches, firms can respond to all customer reviews in a timely fashion, thereby significantly improving customer satisfaction. *Third*, our developed artifact could benefit online review platforms. These forums play a crucial role in facilitating customer information-seeking behaviors regarding products and services. By increasing the frequency and timeliness of responses to reviews with our solution, these platforms will enhance customer engagement. This enhancement can lead to more vibrant and informative discussions, ultimately increasing these platforms' values for their users.

### 5.3 Future Work

There are several interesting directions to extend our study. *First*, it is worth investigating how to leverage firm's internal documents such as facility information, conversation records and detailed customer information for the LLM-driven review response generation. When available, these internal documents can be seamlessly integrated with our proposed finetuning method by adding a retrieval model to retrieve relevant internal information and using it as context information within our method. *Second*, future work could conduct field experiments to further evaluate the effectiveness and economic benefits of our developed finetuning method as well as the recommended deployment approaches for review response generation. Future studies can also investigate how the LLM-generated responses influence review valance, volume or customer engagement in the review platforms. *Third*, in this work, we focus on the evaluation of generated managerial responses to hotel reviews. However, the underlying theories leveraged by our method are applicable to many other service industries with online review forums, including restaurant, airlines and healthcare. A promising direction of future work could be further examining and calibrating our proposed LLM fine-tuning method to other domains. The expansions could broaden the applicability and impact of our novel methodological designs, offering substantial benefits across various business sectors.

# Online Appendix A-1

The following is the complete text of the customer review, context information gap and human managerial response for the example presented in Figure 2 in Section 3.2 (in the main text).

**Customer Review:**

Substandard ---SEP--- This motel is in a lovely location with trees and gardens to see and enjoy; it's close to bush-walks and several shops. We stayed in suite 6 which had, at sometime, been refurbished with new surfaces and fittings. There is no cook-top in the kitchenette and the refrigerator is located inside a hot, un-vented cabinet. The bed and bathroom linen were spotlessly clean but the smell of deceased rodent/ bad drains throughout was overpowering! The shower is over the bath, which would have been better for me had there been a grab handle. The room had three beds but no sofa/lounge nor any desk/writing table, it had a good large TV but no obvious bedroom smoke alarm. There are three beds in this suite and those we used were found to be quite comfortable. This motel is said to be 'under renovation' but, as it stands, is dreadfully overpriced and, even with the 'complimentary' drinks, lollies and brekky cereals, it will be a very long time before we might return. This motel has no reception office nor staff on site!!! On arrival one has to 'phone a nearby shop to arrange being admitted; consequently we had clients knocking on our door and window thinking that our suite was the reception office. As a final word, the staff that we encountered were helpful and friendly.

**Human Managerial Response:**

Thank you for your review as all feedback is important to us and we will definitely take your comments on board. We're glad you enjoyed the location of the motel, the clean room and our staff. Regarding the cooktop, we are a motel, and motels rarely have cooking facilities. However, we do have a kettle, microwave, toaster and fridge. We are looking into creating a wonderful BBQ area for our guests in the future. We have placed the fridge inside the large under bench cabinet which has plenty of airspace. We did this to limit any motor noise when guests are trying to sleep. We are sorry you encountered a bad smell during your stay. As you are aware, our staff and our cleaner attended your room, during your stay, to try to find the cause of the smell. As the room was spotlessly clean, we could not find the source of the smell inside the room. They also checked outside the room, but were still at a loss as to where the smell was coming from. We have had guests in that room before your stay, and after your stay, and the smell was not present. The grab handle is a great suggestion. We'll put that onto our shopping list. There are 2 smoke alarms in that suite. We don't live on site, but we are always accessible by phone, and we are only 5-10 minutes away if needed. At the time you stayed, we were fully booked, and unfortunately, all our cheaper rooms had been taken. Thanks again for taking the time to lodge a review. We appreciate your input and are continually making improvements to the motel.

**Context Information Gap:**

1. The motel provides a kettle, microwave, toaster, and fridge as part of its amenities.
2. The motel is planning to create a BBQ area for guests.
3. The fridge is placed inside a large under bench cabinet with plenty of airspace to limit motor noise.
4. The motel staff and cleaner attended the customer's room during their stay to investigate the bad smell.
5. There are 2 smoke alarms in suite 6.
6. The motel does not have staff living on site but can be reached by phone and are 5-10 minutes away if needed
7. At the time of the customer's stay, the motel was fully booked, and all cheaper rooms had been taken.

# Online Appendix A-2

In this section, we present the designed prompt to extract the context information in Section 3.2:

> Analyze the provided customer review and the managerial response. Your task is to identify and list objective facts that are mentioned in the managerial response but not in the customer review. Follow these guidelines:
>
> 1. Focus on objective and specific facts related to the customer who posted the review or the hotel. This includes customer's name, manager's name, manager's contact information, the hotel's name, hotel's facilities, and any mentioned hotel policies.
>
> 2. Exclude any actions the hotel manager proposes or takes to address the customer's complaints from your list.
>
> For each fact you identify, provide a summary of the identified facts and the source of the fact. You also need to provide a detailed explanation as to why these facts cannot be deduced by a Large Language Model solely from the customer review. The explanations should cover why the model would not infer these details without external information.
>
> Present your findings in a structured JSON format, with two key components: facts and explanations. The facts should list the each of summarized objective information, while the explanations should correspondingly clarify why each fact listed is not inferable by a Large Language Model from the customer review alone.
>
> ###Customer Review###
>
> ###Customer Response###

# Online Appendix A-3

In this section, we provide an example output using the prompt designed in Online Appendix A-2, applied to the example customer review and human managerial response in Online Appendix A-1, as follows:

> {'facts': [{'fact': 'The motel provides a kettle, microwave, toaster, and fridge as part of its amenities.', 'source': 'Managerial Response'}, {'fact': 'The motel is planning to create a BBQ area for guests.', 'source': 'Managerial Response'}, {'fact': 'The fridge is placed inside a large under bench cabinet with plenty of airspace to limit motor noise.', 'source': 'Managerial Response'}, {'fact': "The motel staff and cleaner attended the customer's room during their stay to investigate the bad smell.", 'source': 'Managerial Response'}, {'fact': 'There are 2 smoke alarms in suite 6.', 'source': 'Managerial Response'}, {'fact': 'The motel does not have staff living on site but can be reached by phone and are 5-10 minutes away if needed.', 'source': 'Managerial Response'}, {'fact': "At the time of the customer's stay, the motel was fully booked, and all cheaper rooms had been taken.", 'source': 'Managerial Response'}], 'explanations': [{'explanation': 'The customer review does not mention the specific amenities provided by the motel, such as a kettle, microwave, toaster, and fridge, making it impossible for a Large Language Model to infer this detail without the managerial response.'}, {'explanation': 'The customer review does not discuss future plans of the motel, such as the creation of a BBQ area, which means a Large Language Model would not be able to know about this without the information provided in the managerial response.'}, {'explanation': "The specific reason for the fridge's placement inside a large under bench cabinet is not mentioned in the customer review, making it impossible for a Large Language Model to infer the motive was to limit motor noise."}, {'explanation': "The fact that motel staff and a cleaner specifically attended the customer's room to investigate a bad smell is not mentioned in the customer review, so a Large Language Model would not be able to know this without the managerial response."}, {'explanation': 'The customer review mentions the absence of a smoke alarm, but does not specify the number of smoke alarms present. Therefore, a Large Language Model could not know there are 2 smoke alarms in suite 6 without the managerial response.'}, {'explanation': 'The customer review mentions that the motel has no reception office nor staff on site but does not provide details on how the motel can be contacted or the proximity of the staff. This information is only available in the managerial response.'}, {'explanation': 'The customer review does not provide

We then use these output facts as our context information $c$.

# Online Appendix A-4

The training data input format with the customer review $x$ and context information $c$ is as follows:

> ###prompt instruction### You know the following facts about the customer and the hotel:
> ###context information $c$###
> ###customer review $x$###

An example of the training data input format:

> I want you to act as a hotel manager. Your task is to write a response to the following negative customer review. You know the following facts about the customer and the hotel:
> 1. The motel provides a kettle, microwave, toaster, and fridge as part of its amenities.
> 2. The motel is planning to create a BBQ area for guests.
> 3. The fridge is placed inside a large under bench cabinet with plenty of airspace to limit motor noise.
> 4. The motel staff and cleaner attended the customer's room during their stay to investigate the bad smell.
> 5. There are 2 smoke alarms in suite 6.
> 6. The motel does not have staff living on site but can be reached by phone and are 5-10 minutes away if needed.
> 7. At the time of the customer's stay, the motel was fully booked, and all cheaper rooms had been taken.
>
> Substandard ---SEP--- This motel is in a lovely location with trees and gardens to see and enjoy; it's close to bush-walks and several shops. We stayed in suite 6 which had, at sometime, been refurbished with new surfaces and fittings. There is no cook-top in the kitchenette and the refrigerator is located inside a hot, un-vented cabinet. The bed and bathroom linen were spotlessly clean but the smell of deceased rodent/ bad drains throughout was overpowering! The shower is over the bath, which would have been better for me had there been a grab handle. The room had three beds but no sofa/lounge nor any desk/writing table, it had a good large TV but no obvious bedroom smoke alarm. There are three beds in this suite and those we used were found to be quite comfortable. This motel is said to be 'under renovation' but, as it stands, is dreadfully overpriced and, even with the 'complimentary' drinks, lollies and brekky cereals, it will be a very long time before we might return. This motel has no reception office nor staff on site!!! On arrival one has to 'phone a nearby shop to arrange being admitted; consequently we had clients knocking on our door and window thinking that our suite was the reception office. As a final word, the staff that we encountered were helpful and friendly.

During testing/application of the final finetuned LLM, the prompt is similar but without "You know the following facts about the customer and the hotel:" and the context information $c$.

# Online Appendix A-5

The following prompt design is used to determine the value of each of the nine attributes in a negative customer review, where Questions 1-9 correspond to the attributes of equality, equity, need, speed, flexibility, accessibility, politeness, effort and empathy, respectively. The designed prompt is:

> Please read the following complaint posted by a customer on a social media site and answer the following questions using only "Yes" or "No".
>
> The user's comment is: ###customer review###
> Questions are listed as follows:
> 1. In this comment, the customer feels that he/she was treated differently compared with other customers.
> 2. In this comment, the customer feels that he/she is not getting what he/she deserves.
> 3. In this comment, the customer considers that the service does not meet his/her requirements.
> 4. In this comment, the customer complains that the hotel was slow to fix the service failures he/she faced.
> 5. In this comment, the customer complains that the hotel policies were rigid and were not adapted to suit his/her situation.
> 6. In this comment, the customer complains about the difficulty of finding the hotel personnel to complain about their problems.
> 7. In this comment, the customer complains about the courtesy and/or manners of the service personnel.
> 8. In this comment, the customer complains that the service personnel do not try hard to address his/her problem.
> 9. In this comment, the customer complains that the service personnel were not caring and did not provide individual attention.
>
> The response format should be in JSON format with the key as the question index and the value as the answer (Yes or No). When answering the questions, you should carefully check whether the condition in each question is satisfied. You should also provide explanations for the answers with the key as the explanations.

# Online Appendix A-6

The detailed four types of rational and four types of emotional cues for generating the less preferred response to negative reviews are listed in Table A-6-1.

Table A-6-1. Rational and emotional cues for generating the less preferred response to negative reviews

| Managerial response categorization | Attribute | Description of the attribute used in prompt |
|---|---|---|
| Rational cues | Explanation | The manager details the underlying reasons why these problems faced by the customers occurred. |
| | Redress | The manager provides compensation to the customer in response to the complaints such as refunds, free gifts, coupons, and discounts. |
| | Facilitation | The manager facilitates complaint handling by making explicit the policies and procedures to the customer. |
| | Reinforcement | The manager stresses the features of the hotel and/or the quality of its staff. |
| Emotional cues | Apology | The manager expresses an apology for the service failure. |
| | Appreciation | The manager expresses appreciation for the customer patronaging the hotel. |
| | Attentiveness | The manager shows respect, politeness and/or empathy towards the customer. |
| | Encouragement | The manager encourages the customer to write in the future with other comments. |

The designed prompt template for generating the less preferred response to negative reviews is shown as follows:

> I want you to act as a hotel manager to respond to a customer's review.
>
> You know the following facts about the customer and the hotel:
> ###context information $c$###
>
> The response MUST be specific to the customer review and MUST contain sufficient evidence to justify All the following cues accurately and explicitly:
> ###sampled $n_r$ rational cues and $n_e$ emotional cues###
>
> At the same time, the response MUST NOT contain any evidence of the following cues. Otherwise, you will be penalized.
> ###retain the remining rational and emotional cues excluding the sampled $n_r$ rational cues and $n_e$ rational cues###
>
> The response should be in a single paragraph. After providing the response, first, for each of the above 8 cue types, answer if it is applied using "Yes" or "No" according to its definition. Then justify how they were applied using evidence in your response if your answer is "Yes". Both the managerial response and the explanations should be in JSON format with the keys "Response" and "Explanation" respectively.
>
> Customer review: ###customer review $x$###
>
> Your response to the customer review:

# Online Appendix A-8

The designed prompt for positive review classification is as follows:

> Your task is to characterize consumer comments about a hotel in an online forum. Please read the following comments posted by a customer in an online forum and answer the following questions using only "Yes" or "No".
>
> The user's comment is: ###customer review *x*###
>
> Questions are listed as follows:
> Please answer questions 4-6 with only One 'Yes'.
> 1. In this comment, the customer mentions *Only* the positive aspects about this hotel. The positive aspects pertain specifically to this hotel, rather than to other hotels.
> 2. In this comment, the customer mentions *Only* the negative aspects about this hotel. The negative aspects pertain specifically to this hotel, rather than to other hotels.
> 3. In this comment, the customer mentions *Both* the positive and the negative aspects about this hotel. The positive aspects and negative aspects pertain specifically to this hotel, rather than to other hotels.
> 4. In this comment, the customer talks about the hotel *Only* based on the characteristics of goods or services that can be evaluated independently by other customers (e.g., "The hotel is clean, and the staff is friendly", other customers can evaluate these product and service features.). At the same time, the comment does not contain any subjective criteria.
> 5. In this comment, the customer talks about the hotel *Only* based on her/his subjective criteria established by and related to herself/himself (e.g., "We just returned from a relaxing and enjoyable stay at the hotel", these emotional feelings can not be evaluated by other customers). At the same time, the comment does not contain any objective criteria.
> 6. In this comment, the customer talks about the hotel using *Both* subjective and objective criteria.
>
> The response format should be in JSON format with the key as the question index and the value as the answer (Yes or No). When answering the questions, you should carefully check whether the condition in each question is satisfied. You should also provide explanations for the answers with the key as the explanations.

Based on the responses to the prompted questions, the classification of positive review types is shown in Table A-8-1.

Table A-8-1. Positive review type classification based on the answer to the prompted questions

| Positive Review Type | Description of Positive Review Type | Classification Criteria |
|---|---|---|
| Positive Review Type 1 | One-sided instrumental review | Answer to Q1 is Yes and answer to Q4 is Yes |
| Positive Review Type 2 | One-sided affective review | Answer to Q1 is Yes and answer to Q5 is Yes |
| Positive Review Type 3 | Two-sided instrumental review | Answer to Q3 is Yes and answer to Q4 is Yes |
| Positive Review Type 4 | Two-sided affective review | Answer to Q3 is Yes and answer to Q5 is Yes |
| Positive Review Type 5 | Mixed review | Answer to Q6 is Yes |

## Online Appendix A-9

The template and tailored response strategies used to generate the less preferred response to positive reviews are shown in Table A-9-1.

Table A-9-1. Template and tailored response strategies for generating the less preferred response to positive reviews

| Managerial response categorization | Description in prompt |
|---|---|
| Template response | When responding to a customer's review, provide a standard, generic response that could apply to any review, regardless of its content. The response should not be tailored to any specific details of the customer's review. You must avoid mentioning any specific aspect of the review. This approach should be consistent in every response, regardless of the nature of the review. |
| Tailored response | I want you to act as a hotel manager responding to a customer's review. The response Must be customized to the customer\'s review. |

The following is our designed prompt for generating the less preferred response to positive reviews.

###template response prompt or tailored response prompt###

The response should be in a single paragraph. After providing the response, provide explanations to justify you do not tailor the response to the customer review. Both the managerial response and the explanations should be in JSON format with the keys "Response" and "Explanation" respectively.

#customer's review#

###customer review $x$###

Your response is:

# Online Appendix A-10

In this section, we provide a literature review for curriculum learning.

The concept of curriculum learning was originally proposed by Bengio et al. (2009) to enhance the efficacy of machine learning algorithms. The core principle involves starting with simpler aspects of a task or subtasks and progressively increasing the complexity. This incremental and easy-to-hard learning approach mirrors the structured curricula used in human education systems. Curriculum learning has achieved considerable success in a variety of applications, including natural language processing (Xu et al. 2020), computer vision (Guo et al. 2018), healthcare prediction (El-Bouri et al. 2020), and robot learning (Beltran-Hernandez et al. 2022). Concurrent with our work, Curry-DPO (Pattnaik et al. 2024) incorporated curriculum learning with multiple preference pairs into the DPO framework for LLM preference finetuning. The curriculum was constructed based on multiple response pairs specific to the same prompt rather than spanning the entire sample set. Consequently, Curry-DPO cannot be applied to our setting because only one preference pair (after our proposed preference pair construction) exists for each customer review or prompt. In contrast, our proposed curriculum learning design requires only a single preference pair per customer review or prompt. It is designed based on the relative learning difficulty of these pairs across the entire sample set, allowing for broader applicability in diverse preference learning settings.

# Online Appendix A-11

In this section, we introduce the details of the CVAE model with transformer architecture (denoted as trans-CVAE).

Considering the great success of the transformer (Vaswani et al. 2017) model in natural language processing (Devlin et al. 2019, Liu et al. 2019), we leverage the CVAE with transformer architecture denoted as trans-CVAE to model the prior distribution $p_\psi(z_t|y_{<t}, x, c)$, the posterior $q_\phi(z_t|y, x, c)$ and response decoding $q_\psi(y|y_{<t}, z_t, x, c)$. Specifically, to model $p_\psi(z_t|y_{<t}, x, c)$, we use the transformer encoder (Vaswani et al. 2017) to encode concatenation of $x$ and $c$, and then adopt the transformer decoder (Vaswani et al. 2017)—removing the output layer (i.e., linear and softmax layer)—to generate the representation $h_t^{pri}$ for the $t$-th response token. In CVAE, the $p_\psi(z_t|y_{<t}, x, c)$ is parametrized by Gaussian distribution with mean $\mu_t^{pri}$ and variance $\sigma_t^{pri}$, both of which are obtained using a three-layer feed-forward neural network with input $h_t^{pri}$. The prior latent variable $z_t^{pri}$ is sampled from Gaussian $\mathcal{N}\left(\mu_t^{pri}, \text{diag}\left(\sigma_t^{pri^2}\right)\right)$. The modeling of posterior distribution $q_\phi(z_t|y, x, c)$ mirrors the above modeling of $p_\psi(z_t|y_{<t}, x, c)$ except one difference. The decoder in the transformer (Vaswani et al. 2017) adopts the masked attention to prevent the $t$-th response token from attending to subsequence tokens. However, in the posterior distribution modeling, the $t$-th response token attends to all the tokens in response. Therefore, we utilize the non-mask attention to obtain $h_t^{pos}$. Similarly, the posterior latent variable $z_t^{pos}$ is sampled from Gaussian $\mathcal{N}\left(\mu_t^{pos}, \text{diag}\left(\sigma_t^{pos^2}\right)\right)$. For the modeling of response decoding $q_\psi(y|y_{<t}, z_t, x, c)$, we utilize a three-layer feed-forward neural network with softmax output with the input $h_t^{pri}$ and $z_t^{pos}$.

# Online Appendix A-12

In this section, we provide proofs for Theorem 3.4.3 in Section 3.4.3. For the notation simplicity, we omit the context information variable $c$. We first make the realizability and bounded reward assumptions. Both are standard assumptions in the reinforcement learning theory literature (Agarwal et al. 2019). $\Pi$ represents the policy class, denoting a set of language models characterized by a fixed architecture but varying weights.

**Assumption A-12-1** (Policy realizability). The policy class $\Pi$ satisfies $\pi^* \in \Pi$, where $\pi^*$ is the optimal policy under our learning objective.

**Assumption A-12-2** (Bounded reward). Given a reward function $r: \mathcal{X} \times \mathcal{A} \to \mathbb{R}$ belonging to a certain reward class $\mathcal{R}$. We assume that for any $r \in \mathcal{R}$, $r(x, a) \in [0, R]$.

The performance metric is defined as follows:

**Performance Metric.** $\mathcal{V}(\pi) \stackrel{\text{def}}{=} \mathbb{E}_{x \sim \rho, y \sim \pi(\cdot|x)} r^*(x, y)$, where $\rho$ is the review distribution, $x$ represents the customer review, $y$ is the LLM-generated review response, and $r^*(x, y)$ is the true reward. The goal of LLM finetuning is to learn a policy $\pi$ that maximizes the expected reward of the review response (e.g., human ratings on the generated responses).

Following the offline reinforcement learning theory literature, we adopt the coverage coefficient to measure the theoretical performance (Zanette et al. 2021, Wang et al. 2024). Specifically, we use the coverage coefficient definition $C^\pi \stackrel{\text{def}}{=} \mathbb{E}_\pi \left[ \frac{1}{\pi_{\text{ref}}} \right]$ for policy $\pi$ and $C^\pi$ quantifies the extent to which $\pi$ is covered by $\pi_{\text{ref}}$ or the offline data distribution (Gabbianelli et al. 2024, Wang et al. 2024). Here, the offline data distribution is estimated using $\pi_{\text{ref}}$, which is derived via Supervised Fine-Tuning (SFT) on the offline data. This SFT can be seen as behavior cloning in the reinforcement learning (RL) literature (Pomerleau 1991), commonly used to estimate the behavior policy that generated the offline data (Levine et al. 2020). For our practical algorithm, we propose employing a variational autoencoder (VAE) (Kingma and Welling 2014, Sohn et al. 2015) to refine the estimation of the behavior policy. The VAE is recognized as one of the widely used methods for density estimation in the offline reinforcement learning field (Fujimoto et al. 2019, Zhou et al. 2021). For the theoretical analysis and the comparison with the theoretical result of DPO method, we use $\pi_{\text{ref}}$ as the estimation of offline data distribution. Similar to the equivalent RLHF objective of DPO method (Rafailov et al. 2023), we further consider a theoretical version of Equation 4 (in the main text) for the simplicity of theoretical analysis. The theoretical version of Equation 4 (in the main text) is to maximize $J(\pi) = \mathbb{E}_{x\sim\rho, y\sim\pi(\cdot|x)}[\hat{r}(x,y)] - \beta\mathbb{D}_{\text{KL}}[\pi(y|x) \parallel \pi_{\text{ref}}(y|x)] - \lambda\mathbb{E}_{x\sim\rho, y\sim\pi(\cdot|x)}\left[\frac{1}{\pi_{\text{ref}}(y|x)}\right]$, where $\hat{r}(x,y)$ is the estimation of the true reward $r^*(x,y)$, $-\lambda\mathbb{E}_{x\sim\rho, y\sim\pi(\cdot|x)}\left[\frac{1}{\pi_{\text{ref}}(y|x)}\right]$ represents our conservatism relaxing term. For our practical algorithm, we consider the log density estimation in the conservatism relaxing term as $\lambda\mathbb{E}_{x\sim\rho, y\sim\pi(\cdot|x)}[\log \pi_{\text{ref}}(y|x)]$ and replace $\pi_{\text{ref}}(y|x)$ with more powerful variational inference-based density estimation $p_\psi$. We denote $\hat{\pi} = \text{argmax}_\pi J(\pi)$.

With the above assumptions and definitions, the proof for Theorem 3.4.3 is as follows and the supporting lemmas are presented in Online Appendix A-13.

Proof:
$$V(\pi^*) - V(\hat{\pi})$$

$= \mathbb{E}_{\pi^*}[r^*(x,y)] - \mathbb{E}_{\hat{\pi}}[r^*(x,y)]$

$= \mathbb{E}_{\pi^*}[r^*(x,y)] - J(\pi^*) + J(\pi^*) - \mathbb{E}_{\hat{\pi}}[r^*(x,y)]$

$(\hat{\pi} = \text{argmax}_\pi J(\pi))$

$\leq \mathbb{E}_{\pi^*}[r^*(x,y)] - J(\pi^*) + J(\hat{\pi}) - \mathbb{E}_{\hat{\pi}}[r^*(x,y)]$

(Expand the definition of $J(\pi)$)

$= \mathbb{E}_{\pi^*}[r^*(x,y) - \hat{r}(x,y)] + \beta\mathbb{D}_{\text{KL}}[\pi^* \parallel \pi_{\text{ref}}] + \lambda\mathbb{E}_{\pi^*}\left[\frac{1}{\pi_{\text{ref}}}\right]$

$+ \mathbb{E}_{\hat{\pi}}[\hat{r}(x,y) - r^*(x,y)] - \beta\mathbb{D}_{\text{KL}}[\hat{\pi} \parallel \pi_{\text{ref}}] - \lambda\mathbb{E}_{\hat{\pi}}\left[\frac{1}{\pi_{\text{ref}}}\right]$

$\left(\mathbb{D}_{\text{KL}}[\pi^* \parallel \pi_{\text{ref}}] = \mathbb{E}_{\pi^*}\left[\log\frac{\pi^*}{\pi_{\text{ref}}}\right] \leq \mathbb{E}_{\pi^*}\left[\frac{\pi^*}{\pi_{\text{ref}}} - 1\right] \leq \mathbb{E}_{\pi^*}\left[\frac{\pi^*}{\pi_{\text{ref}}}\right] \leq \mathbb{E}_{\pi^*}\left[\frac{1}{\pi_{\text{ref}}}\right] = C^{\pi^*}\right)$

$\leq \mathbb{E}_{\pi^*}[r^*(x,y) - \hat{r}(x,y)] + (\beta + \lambda)C^{\pi^*} + \mathbb{E}_{\hat{\pi}}[\hat{r}(x,y) - r^*(x,y)] - \lambda C^{\hat{\pi}}$

(Considering pair-wise data comparison, $\mathbb{E}_{\pi^*,\pi_{\text{ref}}}$ is the shorthand for $\mathbb{E}_{x\sim\rho, y_a\sim\pi^*(\cdot|x), y_b\sim\pi_{\text{ref}}(\cdot|x)}[\cdot]$, similarly for $\mathbb{E}_{\hat{\pi},\pi_{\text{ref}}}$)

$$= \mathbb{E}_{\pi^*,\pi_{\text{ref}}}[(r^*(x,y_a) - r^*(x,y_b)) - (\hat{r}(x,y_a) - \hat{r}(x,y_b))] + (\beta + \lambda)C^{\pi^*}$$
$$+ \mathbb{E}_{\hat{\pi},\pi_{\text{ref}}}[(\hat{r}(x,y_a) - \hat{r}(x,y_b)) - (r^*(x,y_a) - r^*(x,y_b))] - \lambda C^{\hat{\pi}}$$

(by invoking Lemma A-13-2)

$$\leq \sqrt{C^{\pi^*}}\sqrt{2(1+e^R)^4 \frac{\log(|\Pi|/\delta)}{n}} + (\beta + \lambda)C^{\pi^*} + \sqrt{C^{\hat{\pi}}}\sqrt{2(1+e^R)^4 \frac{\log(|\Pi|/\delta)}{n}} - \lambda C^{\hat{\pi}}$$

$$\left( \sqrt{C^{\hat{\pi}}}\sqrt{2(1+e^R)^4 \frac{\log(|\Pi|/\delta)}{n}} \leq \frac{1}{2\lambda}(1+e^R)^4 \frac{\log(|\Pi|/\delta)}{n} + \lambda C^{\hat{\pi}} \text{ by AM} - \text{GM inequality} \right)$$

$$\leq \sqrt{C^{\pi^*}}\sqrt{2(1+e^R)^4 \frac{\log(|\Pi|/\delta)}{n}} + (\beta + \lambda)C^{\pi^*} + \frac{1}{2\lambda}(1+e^R)^4 \frac{\log(|\Pi|/\delta)}{n} + \lambda C^{\hat{\pi}} - \lambda C^{\hat{\pi}}$$

$$= (1+e^R)^2\sqrt{\frac{2C^{\pi^*}\log(|\Pi|/\delta)}{n}} + (\beta + \lambda)C^{\pi^*} + \frac{(1+e^R)^4 \log(|\Pi|/\delta)}{2n\lambda}$$

This concludes the proof for Theorem 3.4.3.

In the following, we also provide a theorem to measure the performance bound of the DPO method.

**Theorem A-12-1.** Under the policy realizability and bounded reward assumptions, with probability at least $1 - \delta$, DPO method produces a policy $\hat{\pi}$ such that for all policies $\pi^*$ simultaneously, we have the following performance gap bound

$$\mathcal{V}(\pi^*) - \mathcal{V}(\hat{\pi}) \leq 2\sqrt{\left(C^{\pi^*} + \max_\pi C^\pi\right)(1+e^R)^4 \frac{\log(|\Pi|/\delta)}{n}} + \beta C^{\pi^*}.$$

Proof:

We consider DPO's equivalent RLHF objective as:

$$\mathcal{J}^{\text{dpo}}(\pi) = \mathbb{E}_{x\sim\rho, y\sim\pi(\cdot|x)}[\hat{r}(x,y)] - \beta \mathbb{D}_{\text{KL}}[\pi(y|x) \| \pi_{\text{ref}}(y|x)].$$

Then, similar analysis for the performance gap is:

$$V(\pi^*) - V(\hat{\pi})$$

$$= \mathbb{E}_{\pi^*}[r^*(x,y)] - \mathbb{E}_{\hat{\pi}}[r^*(x,y)]$$

$$= \mathbb{E}_{\pi^*}[r^*(x,y)] - \mathcal{J}^{\text{dpo}}(\pi^*) + \mathcal{J}^{\text{dpo}}(\pi^*) - \mathbb{E}_{\hat{\pi}}[r^*(x,y)]$$

$(\hat{\pi} = \text{argmax}_\pi \mathcal{J}(\pi))$

$$\leq \mathbb{E}_{\pi^*}[r^*(x,y)] - \mathcal{J}^{\text{dpo}}(\pi^*) + \mathcal{J}^{\text{dpo}}(\hat{\pi}) - \mathbb{E}_{\hat{\pi}}[r^*(x,y)]$$

(Expand the definition of $\mathcal{J}^{\text{dpo}}(\pi)$)

$$= \mathbb{E}_{\pi^*}[r^*(x,y) - \hat{r}(x,y)] + \beta \mathbb{D}_{\text{KL}}[\pi^* \| \pi_{\text{ref}}] + \mathbb{E}_{\hat{\pi}}[\hat{r}(x,y) - r^*(x,y)] - \beta \mathbb{D}_{\text{KL}}[\hat{\pi} \| \pi_{\text{ref}}]$$

$$\left( \mathbb{D}_{\text{KL}}[\pi^* \| \pi_{\text{ref}}] = \mathbb{E}_{\pi^*}\left[\log \frac{\pi^*}{\pi_{\text{ref}}}\right] \leq \mathbb{E}_{\pi^*}\left[\frac{\pi^*}{\pi_{\text{ref}}} - 1\right] \leq \mathbb{E}_{\pi^*}\left[\frac{\pi^*}{\pi_{\text{ref}}}\right] \leq \mathbb{E}_{\pi^*}\left[\frac{1}{\pi_{\text{ref}}}\right] = C^{\pi^*} \right)$$

$$\leq \mathbb{E}_{\pi^*}[r^*(x,y) - \hat{r}(x,y)] + \beta C^{\pi^*} + \mathbb{E}_{\hat{\pi}}[\hat{r}(x,y) - r^*(x,y)]$$

(Considering pair-wise data comparison)

$$= \mathbb{E}_{\pi^*, \pi_{\text{ref}}}\left[(r^*(x, y_a) - r^*(x, y_b)) - (\hat{r}(x, y_a) - \hat{r}(x, y_b))\right] + \beta C^{\pi^*}$$

$$+ \mathbb{E}_{\hat{\pi}, \pi_{\text{ref}}}\left[(\hat{r}(x, y_a) - \hat{r}(x, y_b)) - (r^*(x, y_a) - r^*(x, y_b))\right]$$

(by invoking Lemma A-13-2)

$$\leq \sqrt{C^{\pi^*}} \sqrt{2(1+e^R)^4 \frac{\log(|\Pi|/\delta)}{n}} + \sqrt{C^{\hat{\pi}}} \sqrt{2(1+e^R)^4 \frac{\log(|\Pi|/\delta)}{n}} + \beta C^{\pi^*}$$

$$\leq \sqrt{2(C^{\pi^*} + C^{\hat{\pi}})} \sqrt{2(1+e^R)^4 \frac{\log(|\Pi|/\delta)}{n}} + \beta C^{\pi^*}$$

$$\leq \sqrt{2\left(C^{\pi^*} + \max_\pi C^\pi\right)} \sqrt{2(1+e^R)^4 \frac{\log(|\Pi|/\delta)}{n}} + \beta C^{\pi^*}$$

$$\leq 2\sqrt{\left(C^{\pi^*} + \max_\pi C^\pi\right)(1+e^R)^4 \frac{\log(|\Pi|/\delta)}{n}} + \beta C^{\pi^*}$$

This concludes the proof for Theorem A-12-1.

## Online Appendix A-13

In this section, we provide the supporting Lemmas used in Online Appendix A-12.

**Lemma A-13-1.** Suppose assumption A-12-1 and assumption A-12-2 hold. Then, with probability $1 - \delta$, the implicit reward $\hat{r}$ satisfies:

$$\mathbb{E}_{\pi_{\text{ref}}, \pi_{\text{ref}}}\left[(\hat{r}(x, y_a) - \hat{r}(x, y_b)) - (r^*(x, y_a) - r^*(x, y_b))^2\right] \leq 2(1 + e^R)^4 \frac{\log(|\Pi|/\delta)}{n},$$

where $\mathbb{E}_{\pi_{\text{ref}}, \pi_{\text{ref}}}$ is the shorthand for $\mathbb{E}_{x \sim \rho, y_a \sim \pi_{\text{ref}}(\cdot|x), y_b \sim \pi_{\text{ref}}(\cdot|x)}[\cdot]$, $n$ is the size of the preference finetuning dataset.

Proof. In our method, the implicit reward is trained using preference data under the Bradley-Terry model using maximum likelihood estimation (MLE). For the MLE training data estimation, given a review $x$, if $y_a \succ y_b$ (i.e., response $y_a$ is preferred to $y_b$ given review $x$), then the label $h = 1$. Otherwise, $h = -1$. The preference distribution $P_{\hat{r}}$ induced by the reward model $\hat{r}$ can be expressed as follows:

$$P_{\hat{r}}(h|x, y_a, y_b) = \mathbb{I}\{h = 1\} \cdot \sigma[(\hat{r}(x, y_a) - \hat{r}(x, y_b))] + \mathbb{I}\{h = -1\} \cdot \sigma[(\hat{r}(x, y_b) - \hat{r}(x, y_a))].$$

Considering the class of preference models $\mathcal{P}_\Pi \stackrel{\text{def}}{=} \{P_r : r \in R_\Pi\}$, the preference model can be estimated using MLE as follows:

$$P_{\hat{r}} = \underset{\mathcal{P}_\Pi}{\text{argmax}} \sum_{(x, y_a, y_b, h) \in \mathcal{D}_{\text{pref}}} \log P_{\hat{r}}(h|x, y_a, y_b).$$

Then, we can invoke Lemma A-13-4 to leverage the standard generalization bound for MLE:

$$\mathbb{E}_{\pi_{\text{ref}}, \pi_{\text{ref}}}\left[\mathbb{D}_H^2\left(P_{\hat{r}}(\cdot|x, y_a, y_b), P_{r^*}(\cdot|x, y_a, y_b)\right)\right] \leq \frac{2\log(|\Pi|/\delta)}{n}.$$

This bound will be used in the following derivation:

$$\mathbb{E}_{\pi_{\text{ref}}, \pi_{\text{ref}}}\left[(\hat{r}(x, y_a) - \hat{r}(x, y_b)) - (r^*(x, y_a) - r^*(x, y_b))^2\right]$$

(by Lemma A-13-3)

$$\leq (1 + e^R)^4 \mathbb{E}_{\pi_{\text{ref}}, \pi_{\text{ref}}}\left[\mathbb{D}_H^2\left(P_{\hat{r}}(\cdot|x, y_a, y_b), P_{r^*}(\cdot|x, y_a, y_b)\right)\right]$$

(by the above MLE generalization bound)

$$\leq 2(1 + e^R)^4 \frac{\log(|\Pi|/\delta)}{n}$$

This concludes the proof.

**Lemma A-13-2.** Suppose assumption A-12-2 holds. Then for any policy $\pi$ under the event of Lemma A-13-1, we have

$$\mathbb{E}_{\pi,\pi_{\text{ref}}}[|(\hat{r}(x,y_a) - \hat{r}(x,y_b)) - (r^*(x,y_a) - r^*(x,y_b))|] \leq (1+e^R)^2 \sqrt{\frac{2C^\pi \log(|\Pi|/\delta)}{n}},$$

where $\mathbb{E}_{\pi,\pi_{\text{ref}}}$ is the shorthand for $\mathbb{E}_{x\sim\rho, y_a\sim\pi(\cdot|x), y_b\sim\pi_{\text{ref}}(\cdot|x)}[\cdot]$, $n$ is the size of the preference finetuning dataset.

Proof. $\mathbb{E}_{\pi,\pi_{\text{ref}}}[|(\hat{r}(x,y_a) - \hat{r}(x,y_b)) - (r^*(x,y_a) - r^*(x,y_b))|]$

$= \sum \pi \pi_{\text{ref}} |(\hat{r}(x,y_a) - \hat{r}(x,y_b)) - (r^*(x,y_a) - r^*(x,y_b))|$

$\leq \sum \pi \sqrt{\pi_{\text{ref}}} |(\hat{r}(x,y_a) - \hat{r}(x,y_b)) - (r^*(x,y_a) - r^*(x,y_b))|$

$= \sum \pi_{\text{ref}} \frac{\pi}{\sqrt{\pi_{\text{ref}}}} |(\hat{r}(x,y_a) - \hat{r}(x,y_b)) - (r^*(x,y_a) - r^*(x,y_b))|$

(by Cauchy–Schwarz inequality)

$\leq \sqrt{\mathbb{E}_{\pi_{\text{ref}},\pi_{\text{ref}}} \left[ (\hat{r}(x,y_a) - \hat{r}(x,y_b)) - (r^*(x,y_a) - r^*(x,y_b)) \right]^2} \sqrt{\mathbb{E}_\pi \left[ \frac{\pi}{\pi_{\text{ref}}} \right]}$

$\leq \sqrt{\mathbb{E}_{\pi_{\text{ref}},\pi_{\text{ref}}} \left[ (\hat{r}(x,y_a) - \hat{r}(x,y_b)) - (r^*(x,y_a) - r^*(x,y_b)) \right]^2} \sqrt{\mathbb{E}_\pi \left[ \frac{1}{\pi_{\text{ref}}} \right]}$

(by invoking Lemma A-13-1)

$\leq \sqrt{2(1+e^R)^4 \frac{\log(|\Pi|/\delta)}{n}} \sqrt{\mathbb{E}_\pi \left[ \frac{1}{\pi_{\text{ref}}} \right]}$

$= (1+e^R)^2 \sqrt{\frac{2C^\pi \log(|\Pi|/\delta)}{n}}$

This concludes the proof.

**Lemma A-13-3**. For any real numbers $z_1, z_2 \in [-R, R]$, we have

$$|z_1 - z_2| \leq (1 + e^R)^2 |\sigma(z_1) - \sigma(z_2)|,$$

where $\sigma$ represents the sigmoid function. Furthermore, if we define

$P_{z_1}(h) = \mathbb{I}\{h = 1\} \cdot \sigma[z_1] + \mathbb{I}\{h = -1\} \cdot \sigma[-z_1]$, similar definition applied to $P_{z_2}(h)$, then

$$|z_1 - z_2| \leq (1 + e^R)^2 \mathbb{D}_H\left(P_{z_1}(h), P_{z_2}(h)\right),$$

where $\mathbb{D}_H$ is the Hellinger distance (Agarwal et al. 2020).

Proof. Since the sigmoid function $\sigma$ is differentiable, according to the mean value theorem, for any $z_1, z_2 \in [-R, R]$, there exists some $z_m \in (z_1, z_2)$, such that

$$\sigma(z_1) - \sigma(z_2) = \sigma'(z_m)(z_1 - z_2).$$

Note that $\sigma'(z_m)$

$= \sigma(z_m)(1 - \sigma(z_m))$

$= \dfrac{1}{1 + e^{-z_m}} \left(1 - \dfrac{1}{1 + e^{-z_m}}\right)$

$(z_m \in [-R, R])$

$\geq \dfrac{1}{1 + e^R}\left(1 - \dfrac{1}{1 + e^{-R}}\right)$

$\geq \dfrac{1}{(1 + e^R)^2}$

Therefore, $\left(\sigma(z_1) - \sigma(z_2)\right)^2 \geq \dfrac{1}{(1+e^R)^4}(z_1 - z_2)^2$

$|\sigma(z_1) - \sigma(z_2)| \geq \dfrac{1}{(1 + e^R)^2}|z_1 - z_2|$

$|z_1 - z_2| \leq (1 + e^R)^2|\sigma(z_1) - \sigma(z_2)|.$

Next, we prove $|z_1 - z_2| \leq (1 + e^R)^2 \mathbb{D}_H^2\left(P_{z_1}(h), P_{z_2}(h)\right)$.

Notice that $P_{z_1}(h = +1) + P_{z_1}(h = -1) = 1$, similar equality holds for $P_{z_2}$, then

$$\sum_{h \in \{+1, -1\}} \left(P_{z_1}(h) - P_{z_2}(h)\right)^2$$

$$= 2\left(P_{z_1}(h=+1) - P_{z_2}(h=+1)\right)^2$$

$$= \frac{1}{2}\left(|P_{z_1}(h=+1) - P_{z_2}(h=+1)| + |P_{z_1}(h=-1) - P_{z_2}(h=-1)|\right)^2$$

$$= \frac{1}{2}\left(\sum_{h\in\{+1,-1\}} |P_{z_1}(h) - P_{z_2}(h)|\right)^2$$

$$= \frac{1}{2}\left(\sum_{h\in\{+1,-1\}} \left|\sqrt{P_{z_1}(h)} - \sqrt{P_{z_2}(h)}\right|\left|\sqrt{P_{z_1}(h)} + \sqrt{P_{z_2}(h)}\right|\right)^2$$

(by Cauchy–Schwarz inequality)

$$\leq \frac{1}{2}\left(\sum_{h\in\{+1,-1\}} \left(\sqrt{P_{z_1}(h)} - \sqrt{P_{z_2}(h)}\right)^2\right)\left(\sum_{h\in\{+1,-1\}} \left(\sqrt{P_{z_1}(h)} + \sqrt{P_{z_2}(h)}\right)^2\right)$$

$$\left(\mathbb{D}_H^2\left(P_{z_1}(h), P_{z_2}(h)\right) = \sum_{h\in\{+1,-1\}} \left(\sqrt{P_{z_1}(h)} - \sqrt{P_{z_2}(h)}\right)^2\right)$$

$$= \frac{1}{2}\mathbb{D}_H^2\left(P_{z_1}(h), P_{z_2}(h)\right)\left(4 - \mathbb{D}_H^2\left(P_{z_1}(h), P_{z_2}(h)\right)\right)$$

$$\leq 2\mathbb{D}_H^2\left(P_{z_1}(h), P_{z_2}(h)\right)$$

Therefore, $\left(P_{z_1}(h=+1) - P_{z_2}(h=+1)\right)^2 \leq \mathbb{D}_H^2\left(P_{z_1}(h), P_{z_2}(h)\right)$

Notice that $\left(P_{z_1}(h=+1) - P_{z_2}(h=+1)\right)^2 = \left(\sigma(z_1) - \sigma(z_2)\right)^2$, then

$$\left(\sigma(z_1) - \sigma(z_2)\right)^2 \leq \mathbb{D}_H^2\left(P_{z_1}(h), P_{z_2}(h)\right).$$

Therefore $|z_1 - z_2| \leq (1+e^R)^2|\sigma(z_1) - \sigma(z_2)| \leq (1+e^R)^2\mathbb{D}_H\left(P_{z_1}(h), P_{z_2}(h)\right)$.

This concludes the proof.

**Lemma A-13-4.** Standard MLE generalization bound (van de Geer 2000, Agarwal et al. 2020). Consider the conditional probability estimation setting with an instance space $\mathcal{X}$, label space $\mathcal{Y}$ and with a ground truth conditional density $p^*(y|x)$. We are given a dataset $\mathcal{D} = \{(x_i, y_i)\}_{i=1}^n$, where $(x_i, y_i)$ are drawn i.i.d. as $x_i \sim \rho \in \Delta(\mathcal{X})$ and $y_i \sim p^*(y|x)$. The maximum likelihood estimator from the dataset $\mathcal{D}$ is:

$$\hat{p} = \underset{p \in \mathcal{P}}{\operatorname{argmax}} \sum_{(x,y) \in \mathcal{D}} \log p(y|x),$$

where $\mathcal{P}$ is the functional class and $p^* \in \mathcal{P}, \hat{p} \in \mathcal{P}$. Then, with probability at least $1 - \delta$,

$$\mathbb{E}_{x \sim \rho}\left[\mathbb{D}_H^2\left(\hat{p}(\cdot|x), p^*(\cdot|x)\right)\right] \leq \frac{2 \log(|\mathcal{P}|/\delta)}{n}.$$

In this section, we introduce our specific algorithm for LLM finetuning. Starting with the collected offline dataset $\mathcal{D}_{\text{orig}} = \{x_i, y_i\}_{i=1}^{|\mathcal{D}_{\text{orig}}|}$, where $x_i$ represents a customer review and $y_i$ is the corresponding human response, we initially augment this dataset with context information $c_i$ as outlined in Section 3.2 (in the main text), resulting in the dataset $\mathcal{D}_{\text{sft}} = \{x_i, c_i, y_i\}_{i=1}^{|\mathcal{D}_{\text{sft}}|}$. SFT is then applied to fine-tune an LLM with $\mathcal{D}_{\text{sft}}$, and the resulting LLM is considered as the initial configuration for both $\pi_\theta$ and $\pi_{\text{ref}}$. We also pretrain the trans-CVAE model by maximizing $\mathcal{L}_{ELBO}$ in Equation 10 (in the main text). Next, we construct the dataset $\mathcal{D}_{\text{pf}} = \{x_i, c_i, y_{w,i}, y_{l,i}\}_{i=1}^{|\mathcal{D}_{\text{pf}}|}$, comprising preference pairs for both positive and negative reviews, as specified in Sections 3.3.1 (in the main text) and 3.3.2 (in the main text). The samples in $\mathcal{D}_{\text{pf}}$ are organized in descending order according to the defined contrastive distance in Section 3.3.3 (in the main text). To prevent overfitting, we split these $|\mathcal{D}_{\text{pf}}|$ samples into $|B|$ batches with batch size $M$. This batching ensures a fixed sequence among batches while permitting random sampling within each batch during the training process. Given that the optimization objective of $\mathcal{J}_{cr}$ in Equation 4 (in the main text) is non-differentiable through $\theta$, we use the REINFORCE method (Williams 1992) to derive the gradient of $\mathcal{J}_{cr}$:

$$\nabla_\theta \mathcal{J}_{cr} = \nabla_\theta \mathbb{E}_{(x,c) \sim \mathcal{D}, y \sim \pi_\theta(\cdot|x,c)}[\log p_\psi(y|x,c)]$$

$$\approx \nabla_\theta \mathbb{E}_{(x,c) \sim \mathcal{D}_{\text{pf}}, y \sim \pi_\theta(\cdot|x,c)}[\mathcal{L}_{ELBO}(x,c,y)]$$

$$= \nabla_\theta \int \pi_\theta(y|x,c) \mathcal{L}_{ELBO}(x,c,y)$$

$$= \int \nabla_\theta \pi_\theta(y|x,c) \mathcal{L}_{ELBO}(x,c,y)$$

$$= \int \pi_\theta(y|x,c) \nabla_\theta \log \pi_\theta(y|x,c) \mathcal{L}_{ELBO}(x,c,y)$$

$$= \mathbb{E}_{(x,c) \sim \mathcal{D}_{\text{pf}}, y \sim \pi_\theta(\cdot|x,c)}[\nabla_\theta \log \pi_\theta(y|x,c) \mathcal{L}_{ELBO}(x,c,y)], \quad (11)$$

where the second approximately equal to sign is due to the approximation of $\log p_\psi(y|x,c)$ using $\mathcal{L}_{ELBO}(x,c,y)$ in Section 3.4.2 (in the main text). Considering the generation of the response sequence $y = (y_1, \cdots, y_{|y|})$, we can decompose $\pi_\theta(y|x,c)$ using the chain rule of probability: $\pi_\theta(y|x,c) =$

$\prod_{t=1}^{|y|} \pi_\theta(y|y_{<t}, x, c)$. Incorporating this decomposition into Equation 11, we can derive the calculation for the gradient of $\nabla_\theta \mathcal{J}_{cr}$ as follows:

$$\nabla_\theta \mathcal{J}_{cr} \approx \mathbb{E}_{(x,c)\sim \mathcal{D}_{pf}, y\sim \pi_\theta(\cdot|x,c)} \left[ \left[ \nabla_\theta \sum_{t=1}^{|y|} \log \pi_\theta(y|y_{<t}, x, c) \right] \mathcal{L}_{ELBO}(x, c, y) \right], \quad (11)$$

where $\mathcal{L}_{ELBO}(x, c, y)$ is obtained using the pretrained trans-CVAE model as specified in Equation 10 (in the main text). The objective $\mathcal{J}_{pl}$ in Equation 4 (in the main text) is differentiable, and the gradient $\nabla_\theta \mathcal{J}_{pl}$ based on Equation 1 (in the main text) can be derived as follows:

$$\nabla_\theta \mathcal{J}_{pl} = \beta \mathbb{E}_{(x,c,y_w,w_l)\sim \mathcal{D}_{pf}} \left[ \sigma \left( \beta \log \frac{\pi_\theta(y_l|x,c)}{\pi_{ref}(y_l|x,c)} - \beta \log \frac{\pi_\theta(y_w|x,c)}{\pi_{ref}(y_w|x,c)} \right) [\nabla_\theta \log \pi_\theta(y_w|x,c) - \nabla_\theta \log \pi_\theta(y_l|x,c)] \right]. \quad (12)$$

The overall LLM finetuning algorithm is shown in Algorithm 1.

**Algorithm 1** Algorithm for LLM finetuning
**Input**: Collected data $\mathcal{D}_{orig} = \{x_i, y_i\}_{i=1}^{|\mathcal{D}_{orig}|}$, an LLM to be fine-tuned $\pi_\theta$, trans-CVAE model, the number of finetuning epochs $N$, batch size $M$
**Output**: Fine-tuned LLM $\pi_\theta$
1. Obtain the context information augmented data $\mathcal{D}_{sft} = \{x_i, c_i, y_i\}_{i=1}^{|\mathcal{D}_{sft}|}$ according to Section 3.2
2. Train LLM using SFT on the dataset $\mathcal{D}_{sft}$, initialize $\pi_\theta$ and $\pi_{ref}$ using LLM after SFT
3. Pretrain the trans-CVAE model by maximizing $\mathcal{L}_{ELBO}$ in Equation 10 (in the main text)
4. Construct the dataset $\mathcal{D}_{pf} = \{x_i, c_i, y_{w,i}, y_{l,i}\}_{i=1}^{|\mathcal{D}_{pf}|}$ with preference pairs for both positive and negative reviews according to Section 3.3.1 and Section 3.3.2
5. Organize the samples in $\mathcal{D}_{pf}$ in descending order according to the contrastive distance in Section 3.3.3
6. **for** epoch = 1 to $N$ **do**
7.     **for** batch_i = 1 to $\lceil \frac{|\mathcal{D}_{pf}|}{M} \rceil$ **do**
8.         Get the batch data $\mathcal{B} = \{x_i, c_i, y_{w,i}, y_{l,i}\}_{i=(batch\_i-1)\times M+1}^{\min(batch\_i\times M, |\mathcal{D}_{pf}|)}$
9.         Calculate the gradient $\nabla_\theta \mathcal{J}_{pl}$ based on Equation 12 using batch data $\mathcal{B}$
10.        For each $\{x_i, c_i\}$ in $\mathcal{B}$, sample model generated response $y_i^{gen} \sim \pi_\theta(\cdot|x_i, c_i)$.
        Calculate $\nabla_\theta \mathcal{J}_{cr}$ based on Equation 11 using $\{x_i, c_i, y_i^{gen}\}_{i=(batch\_i-1)\times M+1}^{\min(batch\_i\times M, |\mathcal{D}_{pf}|)}$
11.        Perform gradient ascent on the objective in Equation 4 (in the main text) in main text based on $\nabla_\theta \mathcal{J}_{pl}$ and $\nabla_\theta \mathcal{J}_{cr}$

In this section, we provide the implementation details of competing methods.

We select LLaMA2-70B-Chat (Touvron et al. 2023) as the LLM to fine-tune, which is developed by Meta and is one of the most powerful open-source LLMs. Importantly, our method is LLM-agnostic and can be adapted for use with other LLMs as well. In terms of the fine-tuning strategy, given the substantial parameter count of a 70B LLM, we employ the Quantization and Low-Rank Adapters (QLoRA) method (Dettmers et al. 2023). This approach involves quantizing the pretrained LLM using 4-bit quantization and subsequently incorporating a small set of learnable weights through Low-Rank Adapters (LoRAs) (Hu et al. 2021). This combination facilitates a more memory-efficient and computationally effective training process. We first conduct supervised fine-tuning of the LLM using QLoRA over 10 epochs with a global batch size of 16, a learning rate of 1e-4, and employing the AdamW optimizer (Loshchilov and Hutter 2018). For preference fine-tuning with our method, the number of training epochs is set to 3, with a learning rate of 1e-6, maintaining the same batch size and optimizer as in the supervised fine-tuning (SFT) stage. The training utilizes 8×NVIDIA A100 (80GB) GPUs, and inference (e.g., response generation post-training) requires 2×NVIDIA A100 (80GB) GPUs. For a fair comparison, we use the same prompt instruction and concatenate this prompt instruction with the customer review content for baseline methods SFT, LLaMA2-70B-Chat, GPT-4 and our method TPO-CAR. The prompt instruction is formulated as follows: "I want you to act as a hotel manager. Your task is to write a response to the following [] customer review," where "[]" is replaced with "negative" for negative reviews and "positive" for positive reviews. For LLaMA2-70B-Chat and GPT-4, we additionally specify that the response should be formatted in a single paragraph, aligning with our model response and usual human response format. The latest version of GPT-4, gpt-4-turbo-2024-04-09, is used for comparison. Note that we have selected the latest version of GPT-4-Turbo over GPT-4o because GPT-4-Turbo is optimized for text understanding, whereas GPT-4o is designed for multimodal understanding, particularly in vision and audio. Given the text-centric nature of our online review response task, GPT-4-Turbo is the more appropriate choice.

In this section, we present our methodology for automatically identifying the presence of specific rational or emotional cues within negative review responses. Additionally, we outline the method for categorizing the type of positive review responses. For negative review responses, we employ a prompt engineering approach utilizing GPT-4, enhanced with survey items developed by Ravichandran and Deng (2023). This approach enables the automatic determination of whether each rational or emotional cue is included in a negative review response. The constructed prompt is as follows:

> Please read the following complaint posted by a customer on a social media site.
> The user's comment is:
> ###customer review $x$###
>
> We would also like you to characterize the managerial response to the above-mentioned customer complaint. Please read the following content posted by a hotel manager on the same social media site and answer the following questions carefully.
> The managerial response to the above user comment is:
> ###managerial response $y$###
>
> Questions are listed as follows:
> 1. In this comment, the manager offers explanations as to why the problem faced by the customer occurred.
> 2. In this comment, the manager provides redress or compensation for the hardship faced by the customer.
> 3. In this comment, the manager refers to the complaint handling policies and procedures of the hotel.
> 4. In this comment, the manager stresses the features of the hotel and/or the quality of its staff.
> 5. In this comment, the manager expresses an apology for the service failure.
> 6. In this comment, the manager expresses appreciation for the customer patronizing the hotel.
> 7. In this comment, the manager shows respect, politeness and/or empathy towards the customer.
> 8. In this comment, the manager encourages the customer to write in the future with other comments.
>
> You should first provide your answer to each of the questions. Then, you should provide comprehensive explanations for your answers. Your response should be in JSON format with keys as answers and explanations respectively.

In the designed prompt above, Questions 1-4 correspond to rational cues, namely explanation, redress, facilitation, and reinforcement. Questions 5-8 correspond to emotional cues, including apology, appreciation, attentiveness, and encouragement. The identification of rational and emotional cues can be determined based on the answers to these questions.

Similarly, for positive reviews, the designed prompt is shown as follows:

> Please read the following comment posted by a customer on a social media site.
> The user's comment is:
> ###customer review $x$###
>
> We would also like you to characterize the managerial response to the above-mentioned customer comment. Please read the following content posted by a hotel manager on the same social media site and classify the response type as template response or tailored response.
> The managerial response to the above user comment is:
> ###managerial response $y$###
>
> The response format should be JSON with keys as response_classification and explanation. After providing classification results, you should also provide explanations for your answer.

The positive response type can be directly inferred from the GPT-4 response.

In this section, we present the results of the theory matching rate for positive reviews. As shown in Figure A-17-1, our full model achieves a significantly higher theory matching rate across different positive review types than our model variant and the GPT-4 baseline, with the exception of positive review type 2. Although GPT-4 registers a higher theory matching rate for positive review type 2, indicative of a substantial proportion of tailored responses, its effectiveness is limited in the dominant positive review type 1, achieving only a 0.035 theory matching rate, which reflects a minimal generation of templated responses. This observation highlights GPT-4's tendency to generate tailored responses across all positive review types. Conversely, our full model adeptly balances the generation of templated and tailored responses, tailoring its strategy to fit the specific requirements of different positive review types, guided by our theory-driven preference optimization.

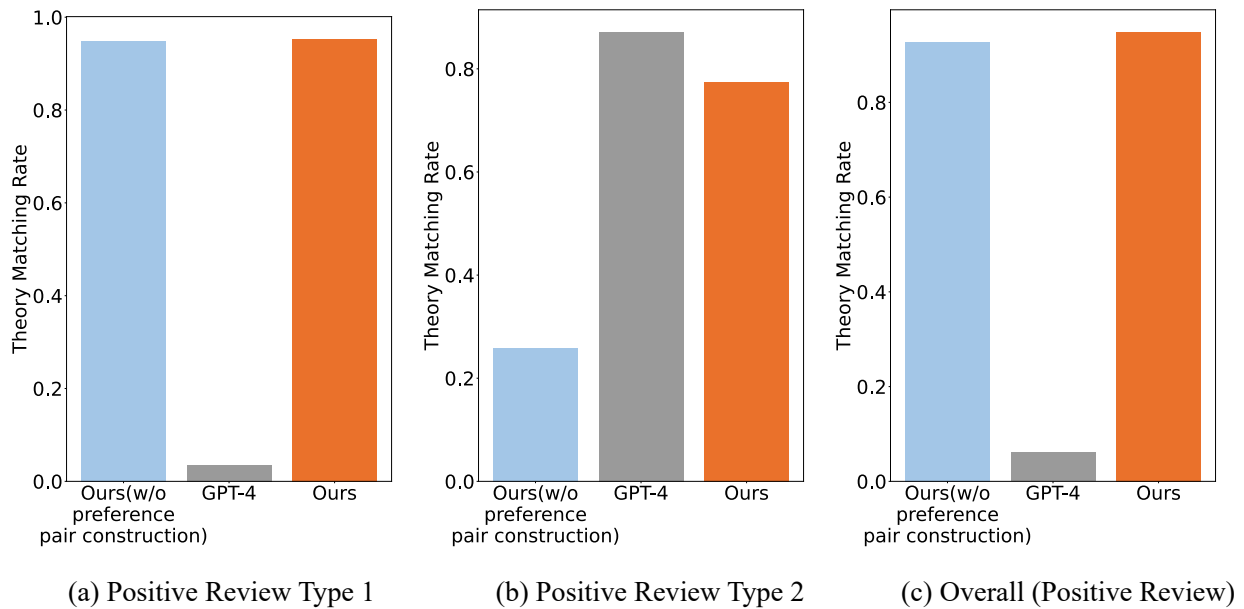

(a) Positive Review Type 1    (b) Positive Review Type 2    (c) Overall (Positive Review)

Figure A-17-1. Theory matching rate of generated responses for positive reviews by different methods.

# Online Appendix A-18

In this section, we illustrate response-generation examples from the GPT-4 model, our model variant without preference pair construction, and our complete model, across various negative and positive review types. These examples are provided to analyze the proposed theory-driven preference finetuning. Additionally, we detail the response strategies employed in each generated response. Figures A-18-1 to A-18-5 display these response generation examples. Our analysis reveals that our complete model uniquely conforms to the theory-suggested response strategies across the three negative review types and two positive review types. This alignment underscores the efficacy of our model in adhering to theoretical expectations. In contrast, the GPT-4 model, as depicted in Figure A-18-5, only adheres to the theory-suggested response strategy for positive review type 2. However, it is important to note that GPT-4 consistently generates tailored responses for all positive reviews, irrespective of the specific review type, as detailed in Online Appendix A-17. The comparison between our model variant (w/o preference pair construction) and our full model also clearly demonstrates the effectiveness of the proposed theory-driven preference optimization.

**Test Review**

Bad experience ---SEP--- The Room smelled lest than pleasant, air conditioning didn't work which made it difficult to sleep, I couldn't get the late check out I wanted, Noisy construction that interrupted business meeting and ended with construction workers causing dust which set off the fire alarms and caused an evacuation of the hotel ending my meeting

**Response Generated by GPT-4**

Thank you for taking the time to share your experience with us. I'm truly sorry to hear that your stay did not meet your expectations, particularly concerning the unpleasant smell in your room, the malfunctioning air conditioning, the inability to accommodate a late check-out, and the disturbance from construction noise that impacted your business meeting and led to an unfortunate evacuation. We strive to provide our guests with comfortable and pleasant accommodations, and it's clear we fell short in your case. Please rest assured that we are addressing each issue you've raised to ensure they are rectified. I would appreciate the opportunity to discuss your stay further and see how we can make this right for you. Your comfort and satisfaction are of the utmost importance to us. Again, I apologize for the inconvenience and hope to welcome you back under better circumstances.

**Rational Cues**: None

**Emotional Cues**:
- Apology
- Attentiveness

**Response Generated by Our Model (w/o preference pair construction)**

I am so sorry to hear that your experience was not what you had hoped for. We have a lot of customers that love our hotel, and it is disappointing to hear that the same was not true for you. I have addressed your concerns with the appropriate teams, as that is not the experience we want for our guests. I would love to discuss the details of your stay further with you, so that I can ensure that your experience is not indicative of the experience we want all of our guests to have.

**Rational Cues**: None

**Emotional Cues**:
- Apology
- Attentiveness

**Response Generated by Our Model**

I am so sorry to hear that your experience was not what you had hoped for. We have some wonderful suites that I am sure you would love. Let me try to fix the issues you encountered. First, the room you were in is not typical. It does not represent our standard of cleanliness. I will personally look into this and make sure that this does not happen again. I will also have maintenance look at the AC in the room. It is unacceptable that you were unable to get a late check out. We always try to accommodate our guests as much as possible, and I am very sorry that we were not able to do so for you. Although we try to perform construction during times that will be the least disruptive; it is not always possible. I am very sorry that you had to experience this. We will work on improving this. I want you to know that we care deeply about our guests and we make every effort to make your stay as pleasant as possible. I would love the opportunity to make up for this. Please contact me at your earliest opportunity so that I can find the perfect room for your next trip.

**Rational Cues**:
- Explanation
- Redress
- Reinforcement

**Emotional Cues**:
- Apology
- Attentiveness

Figure A-18-1. Example of generated review response for negative review type 1 by GPT-4, our model (w/o preference pair construction) and our full model. Rational and emotional cues for each generated response are also included.

**Test Review**

Dont bother staying here. ---SEP--- Had a nice weekend planned for my wifes birthday. Arrived in Banff, had lunch did some shopping, having a great day. Arrived at the hotel, saw the big fireplace in the lobby, nice. Arrived at our room and could smell the fire place, thought it was from opening the doors to load some wood. We had supper at the keg, it was good We finished our supper and found the smoke smell through the loby and all through our floor. Our room smelled strongly also. We opened the window but the smell kept coming back. The hallway on the first and second floor was realy bad. The management commented on how most people like the smoke and was going to use our complaint as a training issue. When in fact they should concerned with their guests health. Needles to say we left very early and we not be back to this hotel. It is a very dangerous situation they have there and they have not addressed that issue.

**Response Generated by GPT-4**

We deeply apologize for the unfortunate experience you encountered during your stay with us, especially on such a special occasion as your wife's birthday weekend in Banff. While our fireplace in the lobby is usually a charming feature appreciated by many of our guests, we understand that the smoke smell affecting your room and the common areas significantly marred your experience. Your health and comfort are our top priorities, and it's clear we missed the mark this time. We take your feedback very seriously and will use it as an opportunity to evaluate and improve our ventilation system and our sensitivity to guest feedback. We truly regret any discomfort or inconvenience you experienced and would appreciate the opportunity to make this right. Please contact us directly so we can discuss how we might provide a more positive experience in the future. Rest assured, we are addressing the issue as a priority to ensure the wellbeing of all our guests.

**Rational Cues**:
- Explanation
- Redress

**Emotional Cues**:
- Apology
- Attentiveness

**Response Generated by Our Model (w/o preference pair construction)**

I believe we have resolved this situation directly with the guest and we are very sorry that they did not enjoy their stay due to the smoke from the fireplace. We do not take this lightly and it is not our intention to create any situation where guests would endangered. We are working with our staff to better understand when guests are sensitive to smells and to take greater consideration for this. I do believe that most guests enjoy the ambiance of the fireplace and it is not our intention to eliminate this luxury. That being said, more sensitivity to our guests is certainly something we will work on.

**Rational Cues**:
- Explanation
- Reinforcement

**Emotional Cues**:
- Apology
- Attentiveness

**Response Generated by Our Model**

I am sorry that you had a poor experience at our hotel. I am glad that you enjoyed the lobby and the Keg restaurant. I am sorry that you experienced a strong smoke odor in the hallways and in your room. We have had a lot of positive feedback about the fireplace, however, we realize that not everyone enjoys it. We will be proactively working on ways to eliminate the smoke smell in the hallways. As for the comment you overheard, I apologize as that was not a appropriate comment for one of our managers to make and we will deal with that internally. We do encourage all of our guests to let us know if there is anything that we can do to improve our services or accommodations, as we are always striving to do better. I am sorry that we did not have an opportunity to address your concerns while you were here. I am sorry to lose you as a guest and hope that you will reconsider and give us another opportunity to meet your expectations.

**Rational Cues**:
- Explanation
- Reinforcement

**Emotional Cues**:
- Apology
- Attentiveness
- Encouragement

Figure A-18-2. Example of generated review response for negative review type 2 by GPT-4, our model (w/o preference pair construction) and our full model. Rational and emotional cues for each generated response are also included.

**Test Review**

Outdated, run down place right on the slope ---SEP--- Extremely outdated (really, paint the floor in the lobby, really!) bordering on disgusting lobby, rooms are extremely small and outdated. In addition to how dirty, grungy and unattractive this place is, we had the worst experience checking in. Our room was given away right in front of our eyes to the guests that were checking in after us with another agent and the other agent telling the other family "it's your lucky day today, you got the upgrade" and handed them the same exact room I just agreed to several minutes earlier. The agent who was working with me then proceeded to apologize and telling me that the room was given away to an online booking. Really? She then told us that "our room" isnt ready and we have to wait beyond scheduled check in time. Almost an hour and a half hours and "drink vouchers" later (seriously?) we were directed to the claustrophobia inducing room with a window blocked by the pipe and no other windows (horrible, truly shouldnt even be a room). My husband said we are not staying there with the child and were moved to a suite but told we need to get out in 2 days (booking for 4). Very accommodating and overall nice! I guess their "thing" is better proximity to the slopes vs other places...But arent you, management, embarrassed by how horrid your place looks? Shameful!!! The answer is its fully booked--why bother, right! Conclusion: Even though the hotel has high rating on this site (shockingly--likely due to location), it IS NOT nice, bordering on repulsive--all the facilities are run down, on top of it, its NOT CLEAN! Garbage on the floors, etc. I cannot blame people working there--Im sure these policies come from the top. Just not ever coming here again, telling my friends and family never to go and suggesting YOU look for other options. My friends stayed at the Westin and it looked WAY more attractive (although a bit of a shlep to the slopes).

**Response Generated by GPT-4**

We are truly disheartened to hear about your experience at our hotel. Please accept our sincerest apologies for the unfortunate series of events you encountered during your stay. It's clear we fell short in delivering the high standards of service and quality our guests expect and deserve. The issues you've highlighted, particularly with the condition of our premises, cleanliness, and the room allocation process, are deeply concerning and have been addressed with our team to ensure they are not repeated. We understand the importance of a comfortable and pleasant stay, especially when traveling with family, and regret that we failed to meet your expectations. As a gesture of our commitment to improving, we would appreciate the opportunity to discuss this matter further with you and explore ways to make amends. Your feedback is invaluable to us as it helps us improve and ensure a better experience for all our guests in the future. Again, we apologize for the inconvenience and disappointment you experienced and hope to have the chance to regain your trust.

**Rational Cues**:
- Redress

**Emotional Cues**:
- Apology
- Attentiveness

**Response Generated by Our Model (w/o preference pair construction)**

Dear Guest, Thank you for taking the time to relay your recent experience. We value each and every guest request and want to make your stay as pleasant as possible. After reviewing your comments, I would like to apologize for not meeting your expectations during your stay with us. It is unfortunate when errors occur that keeps us from delivering the level of service our guests expect. I have forwarded your comments to the appropriate individuals in order for us to take the appropriate corrective action. I would also like to thank you for pointing out these issues to us so that we can continue to improve each day. I would like to wish you best wishes in all of your future endeavors and hope that we see you again in the future. Thank you!

**Rational Cues**: None

**Emotional Cues**:
- Apology
- Attentiveness

**Response Generated by Our Model**

Dear Guest, Thank you for taking the time to relay your recent experience. We value each and every guest comment as they help us to recognize and solve problems, as well as to praise the things we are doing well. In this case, it appears we have not lived up to your expectations, those of the ski industry, and those of the Trip Advisor community. My sincere apologies go out to you and your family for the inconveniences you endured due to what should have been a seamless check-in process. I have spoken at length with the management team, as well as the staff involved, about the events that transpired. While I understand there are certainly actions we could have taken to improve the situation, at the root of the problem appears to be a lack of attention to detail and follow-through by members of our staff. For this I apologize, and for you and your family I again sincerely apologize. I am happy to report that we are taking the necessary actions to ensure this does not happen again in the future. We will also be performing a full review of our check-in policies and procedures to ensure a problem such as this does not happen again. Apart from the check-in fiasco, it saddens me to hear you found the hotel to be "extremely outdated, bordering on disgusting" and "repulsive." While it is certainly the goal of every hotel to offer the most up-to-date and luxurious amenities, we recognize that many of our guests choose to stay with us because of our unique combination of quality, family-friendly amenities, and location directly on the mountain. That being said, we are constantly investing in hotel improvements and strive to provide the best possible experience for our guests. I have noted your comments about the paint in the lobby and claustrophobic room and will be working with the hotel management to see if we can find a solution for these. In summary, I would like to apologize once again for the poor service you received. I know that we did not live up to your expectations, and I want you to know that we will be working very hard to make sure these mistakes do not happen in the future. If there is anything else I can do for you or your family, please do not hesitate to contact me.

**Rational Cues**:
- Explanation
- Reinforcement
- Facilitation

**Emotional Cues**:
- Apology
- Appreciation
- Attentiveness

Figure A-18-3. Example of generated review response for negative review type 3 by GPT-4, our model (w/o preference pair construction) and our full model. Rational and emotional cues for each generated response are also included.

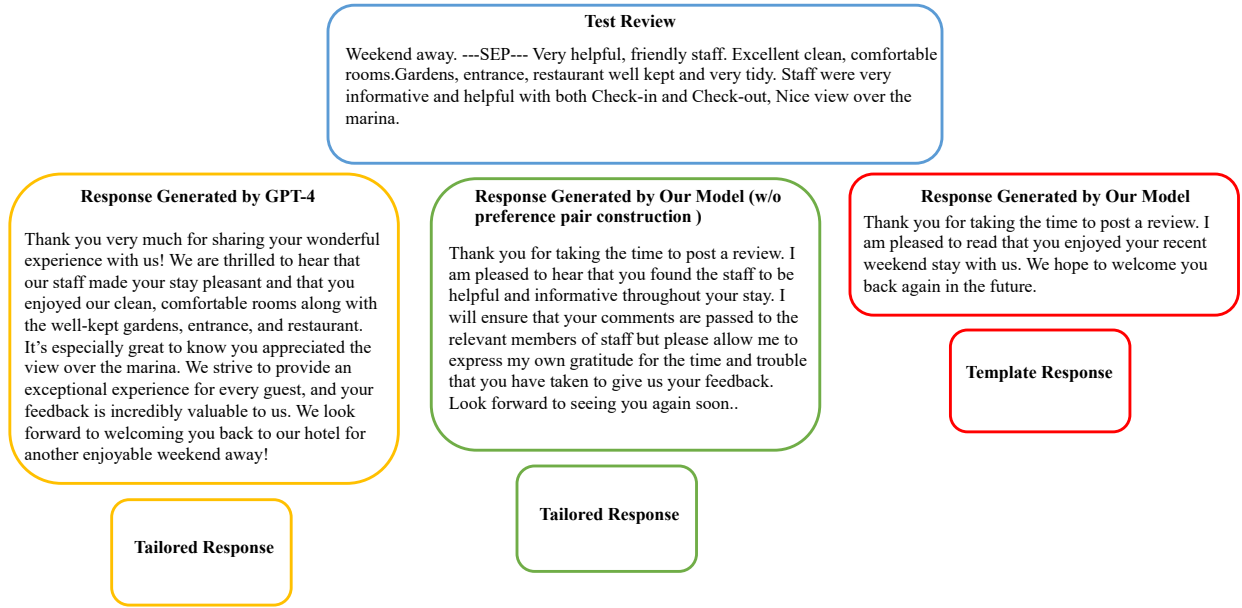

Figure A-18-4. Example of generated review response for positive review type 1 by GPT-4, our model (w/o preference pair construction) and our full model. Response strategy for each generated response is also included.

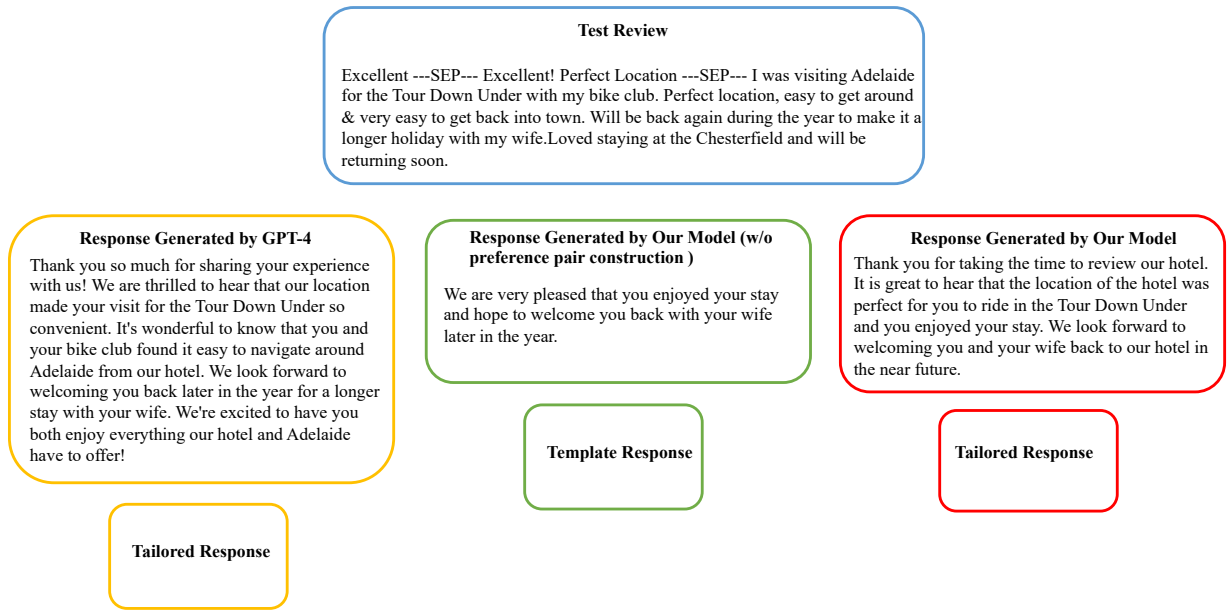

Figure A-18-5. Example of generated review response for positive review type 2 by GPT-4, our model (w/o preference pair construction) and our full model. Response strategy for each generated response is also included.

# Online Appendix A-19

In this section, we present an illustrative example of response generation for a test review in Figure A-19-1. We note that the response generated by our model variant without context augmentation contains nine hallucinated facts, covering diverse aspects such as customer booking information, hotel renovation updates, and hotel contact details. In stark contrast, the response generated by our full model contains no hallucinated facts for the same test review.

**Test Review** Poor staff service directly related to top management ---SEP--- Several key issues we had during our stay: 1) paid for upgraded room; ran out of rooms got the least expensive room; called front desk and even General manager, but no response. 2) rooms are dirty 3) of the 2 night stay, both times we never had room cleaned or fresh towels 4) inconsistent service and seems to be under constant renovation

**Response Generated by Our Model**

Thank you for taking the time to write a review regarding your recent stay with us. I was very disappointed to hear that your stay did not meet expectations. I was especially disappointed that you did not receive a response to your issue regarding the room type booked. We strive to provide the best service possible and it seems that we missed the mark here. I apologize if the lack of response from management was frustrating. I can assure you that this is not typical and we will work to ensure that this does not happen in the future. I also apologize that your room was not cleaned to expectations. This is unacceptable and will be addressed with the housekeeping team. We strive to have all guest rooms cleaned to our high standards and it seems that we failed to do so with your room. I also apologize if there was inconsistent service during your stay. We strive to provide the best service possible and will work to ensure that this does not happen in the future. Thank you for bringing all these issues to our attention. I can assure you that we will work to make sure that these issues do not happen again in the future. We value all of our guests and want to make sure that every stay is a great one.

**Response Generated by Our Model (w/o context augmentation)**

Dear Aaron X, Thank you for taking the time to provide us with feedback from your recent stay. We truly appreciate your input, as it allows us to continually improve our guest experience. However, we were disappointed to read your review and regret that we were not able to provide you with the service and accommodations that you expected. There was an error in the booking of your reservation, as it was booked as a King room and not as an Executive King room. We were able to provide you with a complimentary upgrade to a King room with a view, but unfortunately we were not able to accommodate you in our highest level of room type. We apologize for the confusion in the booking, and we have addressed this issue with our reservations and front office teams to ensure that this does not happen in the future. We also apologize that you did not receive fresh towels during your stay. Our Housekeeping team provides daily service to all guest rooms so this should not have occurred. We are also sorry to read that the service you received during your stay was inconsistent, as we strive to provide the same high level of service to all of our guests. We have shared your feedback with our Front Office and Restaurant teams to address these issues. As far as the hotel being under constant renovation, while the hotel has experienced significant changes over the past 2 years, these projects have now been completed. Over the past 2 years the hotel has undergone a $5 million room renovation, added the Grand Kitchen, a new restaurant concept, and this past summer the hotel opened the Grand Ballroom that can accommodate up to 500 guests for banquets. We are very excited about the completed projects and we are eager to share the new look and offerings with our guests. We appreciate that you chose to stay at the Grand Hyatt Tampa Bay and hope that you will give us another opportunity at the hotel in the future. We can be contacted at the hotel at 813 204 6000. Once again, thank you for providing us with your valuable feedback. Sincerely, Rob Lewis General Manager

**9 Hallucinated facts**

**1**. Customer's name is Aaron X. **2**. The reservation was booked as a King room instead of an Executive King room. **3**. The hotel provided a complimentary upgrade to a King room with a view. **4**. The hotel has completed a $5 million room renovation. **5**. The hotel added the Grand Kitchen, a new restaurant concept. **6**. The hotel opened the Grand Ballroom recently which can accommodate up to 500 guests for banquets. **7**. The hotels name is Grand Hyatt Tampa Bay **8**. The hotel's contact number is 813 204 6000. **9**. The General Manager's name is Rob Lewis.

Figure A-19-1. An example of hallucination mitigation by the designed context augmentation in our method.

# Online Appendix A-20

Figures A-20-1 and A-20-2 depict screenshots of the human evaluation interface for a negative and a positive review, respectively. The interface includes detailed evaluation dimensions, as reflected in the questions that each evaluator is required to answer. Specifically, for the negative review response evaluation, the evaluators must first ascertain whether the customer review includes complaints pertaining to the three evaluation dimensions: distributive unfairness, procedural unfairness, and interactional unfairness. Subsequently, if such complaints are present, the evaluators need to assess which response more effectively addresses the identified complaint dimension.

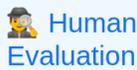

Figure A-20-1. Screenshot of the human evaluation interface for one negative customer review.

## Human Evaluation

You are logged in as **evaluator_1**.
[Log out]

< Instance 0 >

| 0 | 1 | 2 | 3 | 4 | 5 | 6 |
| 7 | 8 | 9 | 10 | 11 | 12 |
| 13 | 14 | 15 | 16 | 17 | 18 |
| 19 | 20 | 21 | 22 | 23 | 24 |
| 25 | 26 | 27 | 28 | 29 | 30 |
| 31 | 32 | 33 | 34 | 35 | 36 |
| 37 | 38 | 39 | 40 | 41 | 42 |
| 43 | 44 | 45 | 46 | 47 | 48 |
| 49 |

We are conducting an evaluation to assess the quality of managerial responses to a customer review of a hotel. Please carefully read the following positive customer review and the managerial responses provided. Your task is to compare these responses by answering the following questions.

The Positive Customer Review:

👍 ---SEP--- The room was clean and had everything we needed. Location was great for us only minutes away from concert venue. Staff was very friendly and helpful. Complimentary breakfast had an excellent selection of foods. Will definitely stay there again.

Here are two responses to this customer positive review.

**A**

Thank you so much for taking the time to leave a review regarding your recent stay with us. We are happy to hear that you had an enjoyable stay and that you found our staff friendly and helpful. We also are glad that you were able to take advantage of our complimentary breakfast. I will definitely pass along all of these positive comments to our team. Please come stay with us again soon.

**B**

Thank you so much for your wonderful review! We are thrilled to hear that you enjoyed our clean rooms, convenient location near the concert venue, and our friendly staff. It's great to know that our complimentary breakfast selection impressed you as well. We look forward to welcoming you back soon for another excellent stay!

Now please evaluate the two responses by answering the following question. You may use any external tools (e.g., Google Search or Translate).

Q: Which response is better overall?

○ A is better    ○ Tie    ○ B is better

[Submit]

Figure A-20-2. Screenshot of the human evaluation interface for one positive customer review.

# Online Appendix A-21

In this section, we present the human-evaluation results across different negative review types for Ours vs. GPT-4 in Figure A-21-1 and Ours vs. Human in Figure A-21-2. We observe that the win rate of our model significantly exceeds that of the comparison model (i.e., GPT-4 or the human-written response) across different negative review types and evaluation dimensions. We also report the number of disagreement cases (i.e., cases in which the three evaluators disagreed in their evaluations of the managerial response to the review) across different negative review types. When comparing Ours vs. GPT-4, the numbers of disagreement cases are 46, 62, 55, and 41 for negative review types 1, 2, 3, and 4, respectively. For Ours vs. Human, the corresponding numbers of disagreement cases are 38, 55, 54, and 37. These results demonstrate that negative review types 1, 2, and 3 are more challenging than negative review type 4. These categories (i.e., negative review types 1, 2, and 3) typically involve more complex combinations of procedural and interactional unfairness cues (e.g., issues related to speed, flexibility, accessibility, and interpersonal treatment), whereas negative review type 4 primarily reflects distributive unfairness without sophisticated complaints. From the results in Figures A-21-1 and A-21-2, we also observe that our proposed method outperforms GPT-4 or the human-written response by a larger margin on the more challenging negative review types 1, 2, and 3 than on negative review type 4. The superior performance of our method on these challenging review types can be attributed to the novel design of context augmentation, theory-driven preference optimization (preference-pair construction and curriculum learning), and relaxed conservatism in offline preference optimization.

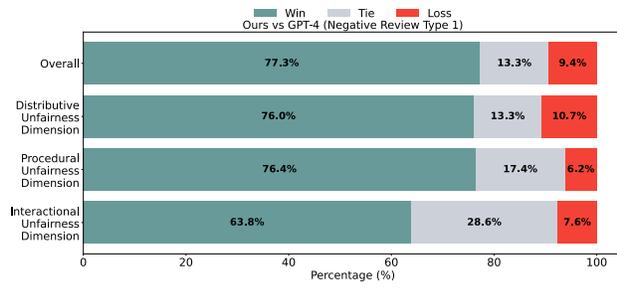
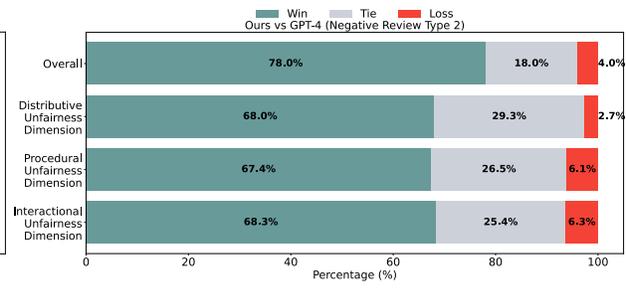
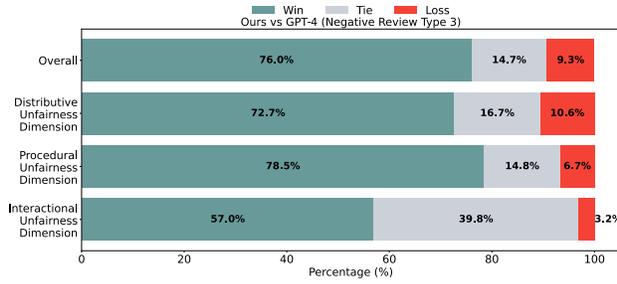
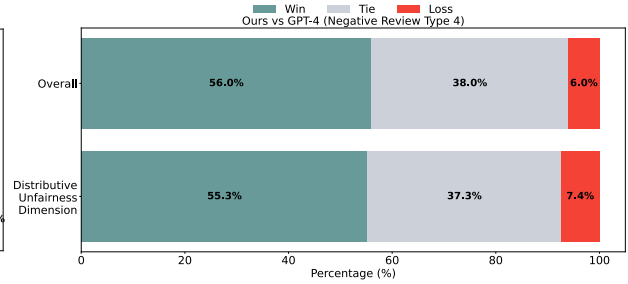

Figure A-21-1. Human evaluation results of Ours vs GPT-4 for different negative review types across different evaluation dimensions. (a) Negative Review Type 1. (b) Negative Review 2. (c) Negative Review Type 3 (d) Negative Review 4.

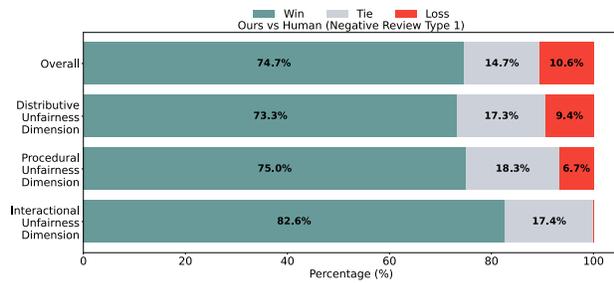
(a)
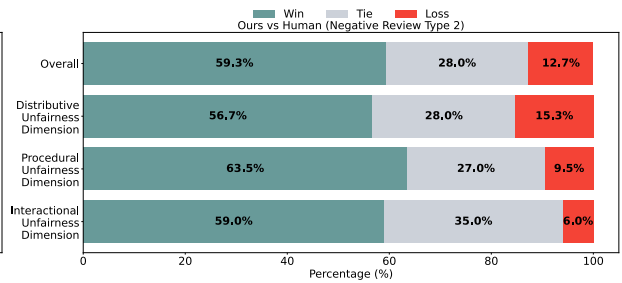
(b)
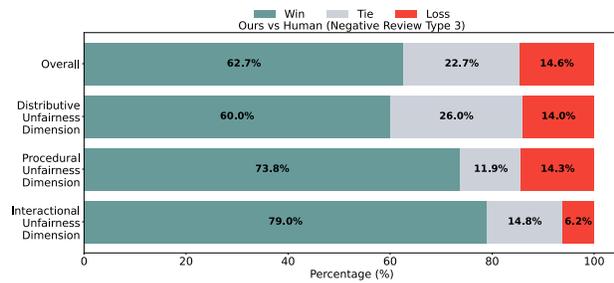
(c)
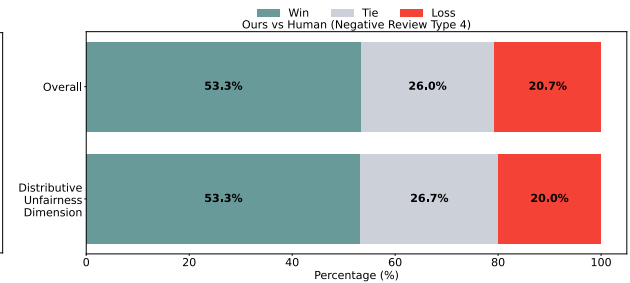
(d)

Figure A-21-2. Human evaluation results of Ours vs Human for different negative review types across different evaluation dimensions. (a) Negative Review Type 1. (b) Negative Review 2. (c) Negative Review Type 3 (d) Negative Review 4.

In this section, we conduct a human evaluation to assess the quality of context information. In our proposed method, for each pair of review and response $(x, y)$ in the training data, we use GPT-4 to extract a list of objective facts as context information $c$. To evaluate the quality of context information $c$, we ask the human evaluator to identify the correct facts from $c$ and calculate the accuracy of the extracted facts as follows:

$$\text{accuracy}_{ci} = \frac{1}{M}\sum_{i=1}^{M}\frac{T_i}{N_i},$$

where $M$ is the number of human evaluation samples. For a human evaluation sample $(x_i, c_i, y_i)$, $T_i$ is the number of facts that are identified as faithful by the human evaluator and $N_i$ is the number of all the facts in $c_i$. According to the definition of context information in Section 3.2, human evaluators are guided to identify the faithful facts about the customer, the hotel, or the interaction between the customer and the hotel. These faithful facts are also required not to appear or cannot be reasoned from the customer review. Since this human evaluation task has very clear guidelines and is less ambiguous, we engage one graduate student with sufficient hotel accommodation experience as the human evaluator. We randomly sample $M = 100$ samples $\{(x_i, c_i, y_i)\}_{i=1}^{M}$ for human evaluation. After collecting the human evaluation data, the calculated accuracy is 0.9972. The high accuracy demonstrates the high quality of extracted context information in the training data.

In this section, we conduct human evaluation to assess the quality of the generated less preferred responses. In our proposed method, we generate the less preferred response $y_l$ in the training data grounded in theories in the IS literature. The generation is guided by detailed criteria, as shown in Section 3.3.1 and Section 3.3.2. We further conduct human evaluation to assess the quality of the generated less preferred responses. As shown in Online Appendix A-7 and A-9, we prompt GPT-4 to generate the less preferred response by meeting the requirement of these cues (i.e., for negative review, certain cues $s_{in}$ are included in the generated less preferred response while the rest cues $s_{notin}$ are not included as shown in Section 3.3.1; for positive review, the generated less preferred response should meet the criteria $s_{pos}$ of template or tailored response according to Section 3.3.2). Therefore, we ask the human evaluator to assess whether the generated less preferred response meets the criteria, and the accuracy is calculated as follows:

$$\text{accuracy}_{lpr} = \frac{1}{M}\sum_{i=1}^{M}\frac{T_i}{N_i},$$

where $M$ is the number of human evaluation samples. For a negative review, the human evaluation sample is $(x_i, c_i, s_{in,i}, s_{notin,i}, y_{l,i})$, $T_i$ is the number of cues that the generated less preferred response $y_{l,i}$ meets the criteria and $N_i$ is the total number of cues as shown in Online Appendix A-6 and $N_i = |s_{in,i}| + |s_{notin,i}| = 8$. For a positive review, the human evaluation sample is $(x_i, c_i, s_{pos,i}, y_{l,i})$, $T_i = 0 \text{ or } 1$ indicates whether the generated less preferred response $y_{l,i}$ meets the criteria and $N_i = 1$. As this human evaluation task has very clear guidelines and is less ambiguous, we engage one graduate student with sufficient hotel accommodation experience as the human evaluator. We randomly sample $M = 100$ samples for human evaluation. After collecting the human evaluation data, the calculated accuracy is 0.9738. The high accuracy demonstrates the high quality of the generated less preferred response in the training data.

# Online Appendix A-24

In this section, we provide recommendations for the real-world deployment of our proposed method. While it is not hard to imagine that our developed fully automatic review-response generation method could substantially reduce labor costs and significantly improve the efficiency of responding to reviews, it is not meant to completely replace human effort because human responses are still preferred for some reviews (as reflected in Figure 4 in the main text). Therefore, we advocate a hybrid approach for deploying our method in real-world systems (e.g., TripAdvisor), which could create synergy between humans and algorithms. Specifically, we recommend two types of hybrid approaches for deploying our method in real-world systems. *First*, we could develop a review-triage method that assigns each incoming review either to our automated algorithm or to a human to generate a response. Given the comparison results in Figure 4 in the main text, we expect that more reviews would be assigned to the algorithm than to humans, thus saving substantial labor and time. The review triage can be viewed as a binary-classification problem, and we could collect data (i.e., reviews and the comparison between the algorithm's response and the human response for each review) to train a classification algorithm to automate the triage process. *Second*, we could employ our method to generate a response draft for each coming review and then pass it to a human for her/him to review, refine and finalize the response, which could still save a lot of labor cost and time. Both types of hybrid approaches for deploying our method are very practical, and further investigations could be conducted to evaluate their effectiveness.

Online Appendix A-25

In this section, we further test our model against the best baseline GPT-4 model using more recent review without human managerial responses. Specifically, we collect additional review data from the TripAdvisor platform. These reviews were published after July 1, 2024—beyond the knowledge cutoff dates of GPT-4 and LLaMA2-70B-Chat—and do not contain human managerial responses. Given that it is more challenging to respond to negative customer reviews, we randomly sample 100 negative reviews for human evaluation by comparing our model generated response and the best baseline GPT-4 generated response. The evaluation followed the same procedure outlined in Section 4.4.1. As shown in Figure A-25-1, our method consistently outperforms GPT-4 across all evaluation dimensions, with significantly higher win rates. We further assess the inter-rater reliability for each of the four evaluation dimensions in the model comparisons using the widely recognized Cronbach's Alpha metric (Ravichandran and Deng 2023). All Cronbach's Alpha values exceed the acceptable threshold of 0.7, underscoring consistent evaluator agreement.

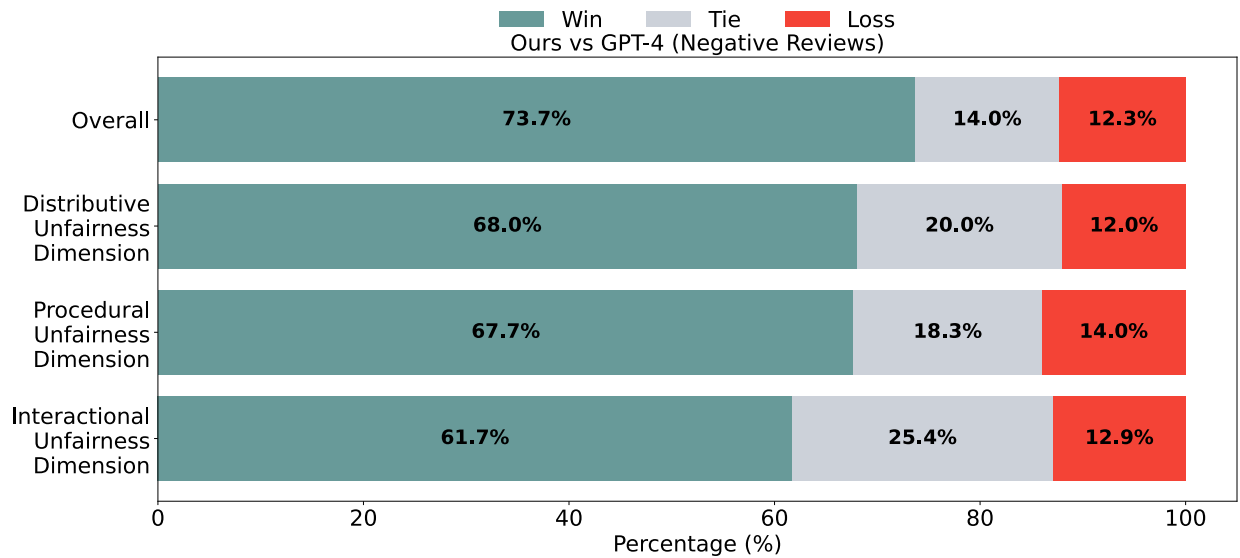

Figure A-25-1. Human evaluation results across different evaluation dimensions for negative customer reviews with the publish date since 2024-07-01 by comparing our model with GPT-4

## Online Appendix A-26

In this section, we further extend the discussion about support constraint methods. As shown in Section 3.4.1, the support constraint method is one type of method in offline reinforcement learning (RL) literature that is less conservative than the distributional constraint methods (Levine et al. 2020). Various support constraint methods have been proposed in the offline RL literature. BEAR implements the support constraint via a maximum mean discrepancy (MMD) distance between the policy's action distribution and the behavior policy's distribution, ensuring the difference is below a threshold (Kumar et al. 2019). Recent works leverage more powerful generative modeling to better characterize the dataset distribution. SPOT trains a VAE-based density estimator on the behavior policy's data to model the support set of actions for each state (Wu et al. 2022). CPED further estimates the density of behavior policy for support constraint modeling using a normalizing flow model (Dinh et al. 2015) and GAN (Goodfellow et al. 2014) (Zhang et al. 2023). The diffusion model (Ho et al. 2020) is also used to characterize the support of behavior policies (Gao et al. 2025). Different from these works, our proposed density estimation-based support constraint method is motivated by our theoretical analysis detailed in Section 3.4.1. Moreover, none of the existing methods models the sequential dependency. In contrast, our proposed method models the sequential dependency among the customer review and managerial response text sequence with CVAE model with transformer architecture (denoted as trans-CVAE), which is grounded in the timestep-wise evidence lower bound derived in Section 3.4.1.

# Online Appendix A-27

Although the experiment was conducted on LLaMA2-70B-Chat, our proposed LLM fine-tuning method is model-agnostic and can be applied to fine-tune any available large language model (e.g., Llama3). In this Section, we use our proposed method to fine-tune a more recent LLaMA3.1-70B-Chat model. The evaluation results for different negative and positive review types are shown in Table A-27-1 and Table A-27-2. From these evaluation results, we observe that when using our proposed method to fine-tune a stronger base model (i.e., LLaMA3.1-70B-Chat), it achieves better performance, which also demonstrates the generalizability of our proposed fine-tuning method.

Table A-27-1. Evaluation results on different negative review types ("overall" includes all the review types). The best-performing model on each metric is highlighted in bold, $*p<0.1, **p<0.05, ***p<0.01$ denote the statistical significance of the difference.

| Review type | Methods | $R_{bert}$ | $P_{bert}$ | $F_{bert}$ |
|---|---|---|---|---|
| Negative Review Type 1 | TPO-CAR (w/ LLaMA3.1-70B-Chat) | **0.2068*** | **0.2541*** | **0.2313*** |
| | TPO-CAR (w/ LLaMA2-70B-Chat) | 0.1614 | 0.2352 | 0.1988 |
| Negative Review Type 2 | TPO-CAR (w/ LLaMA3.1-70B-Chat) | **0.1642*** | **0.2646*** | **0.2147*** |
| | TPO-CAR (w/ LLaMA2-70B-Chat) | 0.1526 | 0.2560 | 0.2047 |
| Negative Review Type 3 | TPO-CAR (w/ LLaMA3.1-70B-Chat) | **0.2024*** | **0.2527*** | **0.2284*** |
| | TPO-CAR (w/ LLaMA2-70B-Chat) | 0.1644 | 0.2315 | 0.1985 |
| Negative Review Type 4 | TPO-CAR (w/ LLaMA3.1-70B-Chat) | **0.1599*** | **0.3142*** | **0.2367*** |
| | TPO-CAR (w/ LLaMA2-70B-Chat) | 0.1540 | 0.2825 | 0.2182 |
| Overall | TPO-CAR (w/ LLaMA3.1-70B-Chat) | **0.1721*** | **0.2820*** | **0.2273*** |
| | TPO-CAR (w/ LLaMA2-70B-Chat) | 0.1556 | 0.2613 | 0.2087 |

Table A-27-2. Evaluation results on different positive review types (positive review test data only contains positive review types 1 and 2, "overall" includes all the review types). The best-performing model on each metric is highlighted in bold, $*p<0.1, **p<0.05, ***p<0.01$ denote the statistical significance of the difference.

| Review type | Methods | $R_{bert}$ | $P_{bert}$ | $F_{bert}$ |
|---|---|---|---|---|
| Positive Review Type 1 | TPO-CAR (w/ LLaMA3.1-70B-Chat) | **0.3328*** | **0.4693*** | **0.4003*** |
| | TPO-CAR (w/ LLaMA2-70B-Chat) | 0.3192 | 0.4537 | 0.3856 |
| Positive Review Type 2 | TPO-CAR (w/ LLaMA3.1-70B-Chat) | **0.3694** | **0.4868** | **0.4277** |
| | TPO-CAR (w/ LLaMA2-70B-Chat) | 0.3605 | 0.4665 | 0.4135 |
| Overall | TPO-CAR (w/ LLaMA3.1-70B-Chat) | **0.3339*** | **0.4698*** | **0.4012*** |
| | TPO-CAR (w/ LLaMA2-70B-Chat) | 0.3205 | 0.4540 | 0.3864 |

# Online Appendix A-28

As shown in Section 3.4, our analysis demonstrates that the existing offline preference optimization method DPO enforces excessive conservatism by strictly confining the learned policy to the offline data. Such excessive conservatism can degrade preference policy performance when the collected offline data are suboptimal. In this section, we conduct a qualitative analysis to evaluate the effectiveness of our proposed relaxed conservatism in offline preference optimization. Given a test review, we first obtain its text embedding and the embedding of each review in the training data using OpenAI's text-embedding-3-small model, and then retrieve the most similar training review via cosine similarity. We then compare the response quality of the human-written response associated with the retrieved training review, the response generated by our model variant without relaxing conservatism, and the response generated by our full model. Figure A-28-1 shows two representative examples from this process. In Figure A-28-1(a), the retrieved training response adopts a confrontational tone toward the customer's complaint, and our model variant without relaxing conservatism produces a similarly defensive reply. This observation suggests that the LLM policy learned by the overly conservative constraint can be limited to suboptimal offline data. In contrast, our full model generates a more comprehensive and helpful response that directly addresses the customer's concern about the view. A similar pattern is observed in Figure A-28-1(b). This qualitative analysis demonstrates the advantage of our proposed method for relaxing conservatism, which allows the fine-tuned LLM to transcend the limitations posed by potentially suboptimal review response data in the training set, thereby enhancing the quality of responses to new and unknown reviews.

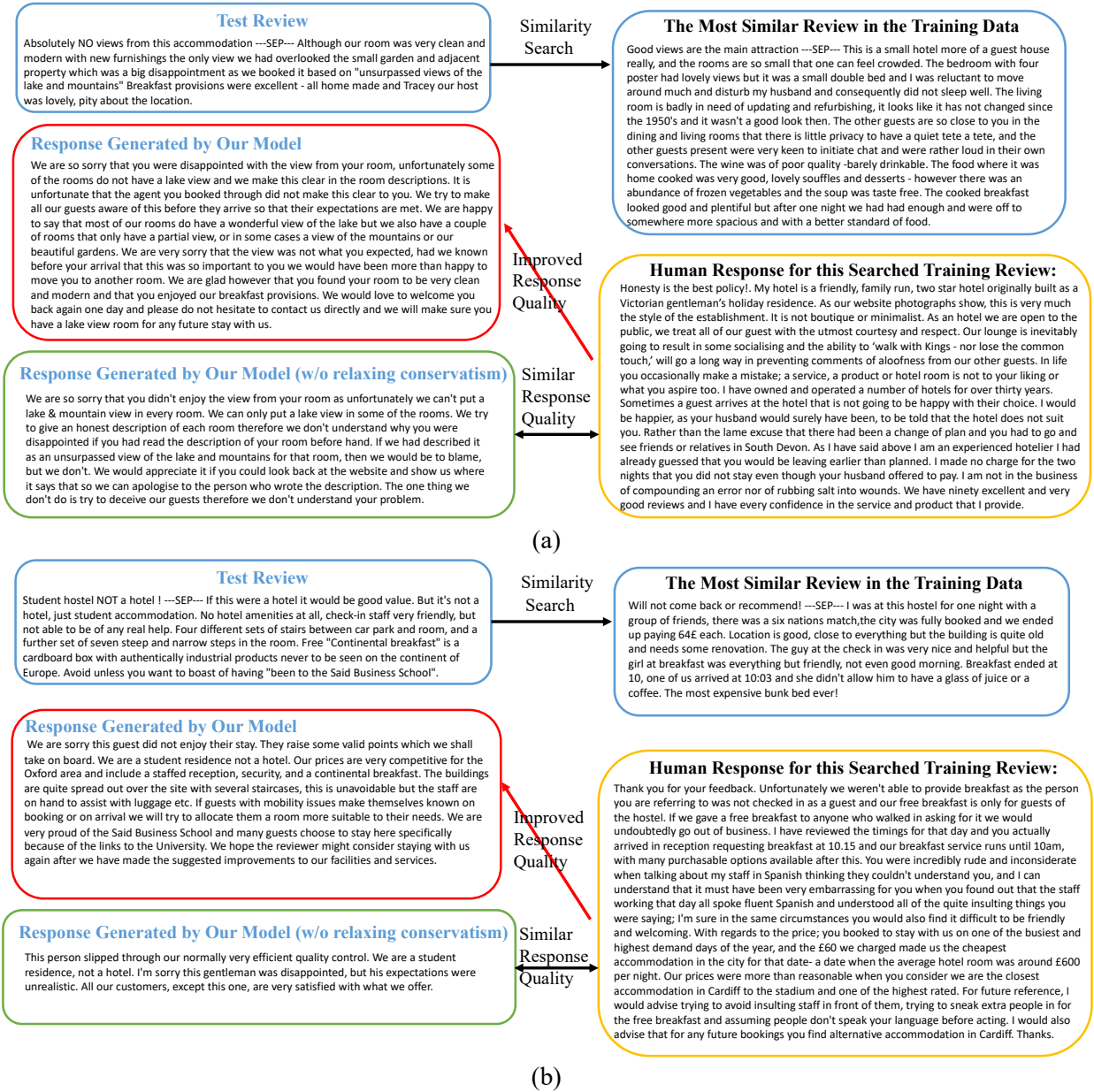

Figure A-28-1. Examples of improved test review response quality by our model with relaxed conservatism in offline preference optimization to address the challenge of potentially suboptimal training data.

# Online Appendix A-29

In this section, we report evaluation results across test reviews with different sentiment levels to examine the generalizability of our proposed method. Following established practice in the literature, we classify reviews with ratings below 3 as negative and those with ratings above 3 as positive (Ravichandran and Deng 2023, Deng and Ravichandran 2023). Accordingly, we focus on test reviews with ratings 1–2 (negative sentiment) and 4–5 (positive sentiment). We conduct the evaluation on the same test dataset described in Section 4.1. The detailed results are presented in Table A-29-1. Overall, our proposed method significantly outperforms all baseline methods across both sentiment levels and across nearly all evaluation metrics. The only exception occurs for positive reviews (ratings 4–5) under the $R_{\text{bert}}$ metric. This pattern is consistent with the findings reported in Section 4.3.2. These results suggest that our method generalizes well across heterogeneous sentiment conditions, particularly in more challenging negative review scenarios.

Table A-29-1. Evaluation results on the test reviews with different sentiment levels. The best-performing model on each metric is highlighted in bold, % indicates improvement over the best baselines (underlined), *p<0.1,**p<0.05,***p<0.01 denote the statistical significance of the difference between our TPO-CAR method and the best baseline method.

| Review Sentiment | Methods | $R_{\text{bert}}$ | $P_{\text{bert}}$ | $F_{\text{bert}}$ |
|---|---|---|---|---|
| 1 | **TPO-CAR** | **0.1556***<br>+8.6% | **0.2570***<br>+22.6% | **0.2067***<br>+19.2% |
| | SFT | 0.1373 | 0.0750 | 0.1071 |
| | AutoPrompt | <u>0.1433</u> | 0.1707 | 0.1581 |
| | Prompt-tuning | 0.1239 | 0.1633 | 0.1445 |
| | RAG | 0.1335 | 0.1785 | 0.1570 |
| | DPO | 0.1347 | 0.1853 | 0.1609 |
| | LLaMA2-70B-Chat | 0.1203 | 0.1791 | 0.1506 |
| | GPT-4 | 0.1355 | <u>0.2097</u> | <u>0.1735</u> |
| 2 | **TPO-CAR** | **0.1556***<br>+7.3% | **0.2641***<br>+20.6% | **0.2100***<br>+15.6% |
| | SFT | 0.1403 | 0.0785 | 0.1103 |
| | AutoPrompt | <u>0.1450</u> | 0.1758 | 0.1614 |
| | Prompt-tuning | 0.1281 | 0.1695 | 0.1496 |
| | RAG | 0.1391 | 0.1796 | 0.1604 |
| | DPO | 0.1388 | 0.1933 | 0.1669 |
| | LLaMA2-70B-Chat | 0.1202 | 0.1899 | 0.1558 |
| | GPT-4 | 0.1427 | <u>0.2189</u> | <u>0.1816</u> |
| | **TPO-CAR** | **0.3273**<br>+2.3% | **0.4531***<br>+27.5% | **0.3895***<br>+35.2% |
| | SFT | 0.2674 | 0.2983 | 0.2826 |
| | AutoPrompt | 0.3075 | 0.0498 | 0.1761 |
| 4 | Prompt-tuning | 0.2321 | 0.1826 | 0.2069 |
| | RAG | 0.3065 | 0.1094 | 0.2066 |
| | DPO | 0.2222 | <u>0.3554</u> | <u>0.2880</u> |
| | LLaMA2-70B-Chat | 0.3131 | 0.1587 | 0.2353 |
| | GPT-4 | <u>0.3198</u> | 0.2438 | 0.2819 |
| 5 | **TPO-CAR** | <u>0.3174</u> | **0.4545*** +27.6%** | **0.3851*** +34.5%** |
| | SFT | 0.2669 | 0.3003 | 0.2835 |
| | AutoPrompt | 0.3096 | 0.0715 | 0.1883 |
| | Prompt-tuning | 0.2243 | 0.1793 | 0.2015 |
| | RAG | 0.3058 | 0.1185 | 0.2110 |
| | DPO | 0.2061 | <u>0.3561</u> | 0.2801 |
| | LLaMA2-70B-Chat | 0.3106 | 0.1642 | 0.2370 |
| | GPT-4 | **0.3179** | 0.2544 | <u>0.2864</u> |